# Advancements in Diabetic Retinopathy Classification Using Novel Multi-Stacked U-Net Architectures.

[1] S. Navaneetha Krishnan, [2]Ameya Uppina, [3]Talluri Krishna Sai Teja, [4]Joe Dhanith P R

[1,2,3,4]School of Computer Science and Engineering, Vellore Institute of Technology Chennai Campus, India

Email: [1]krishnannavaneetha467@gmail.com, [2]ameyauppina2003@gmail.com, [3]teja05443@gmail.com, [4]joedhanith.pr@vit.ac.in

**Abstract:** Diabetes leads to several health complications throughout the body, with the complication in the eyes being Diabetic Retinopathy. It leads to damaging or abnormal changes in the blood vessels' structure, causing vision loss. The need for massive screening of a large population of diabetic patients has created a necessity for computer-aided, fully automatic diagnosis of DR. CNNs have shown great promise in the classification of DR based on the Fundus images. Building a reliable model faces considerable challenges due to the limited availability of high-quality annotated datasets and issues with varying image quality and class imbalances. This paper aims to demonstrate the UNET, and its Stacked Architecture variant and their performance across several datasets. This paper has compared and studied the UNET architecture's performance across multiple stacks and has come to a soft conclusion that the architecture performs its best at three stacks, and any more leads to deterioration of its performance. The model aims at classifying the images into five categories, 0 to 4, where 0 is no DR, and 4 is proliferative DR.

## I. INTRODUCTION

Diabetic Retinopathy (DR) is a leading cause of blindness in developed countries, occurring mainly due to the rise in diabetes among the population. It is one of the very serious side effects of Diabetes Mellitus (DM), which is a condition that damages the retina and causes serious vision impairment. According to multiple research studies, it is estimated that approximately 420 million people globally are currently diagnosed with diabetes.[1][2] The number of people suffering from this chronic condition has doubled over the past three decades, with projections indicating a continuous upward trend. Among those affected, it is anticipated that around one-third will develop diabetic retinopathy (DR). In the United States alone, 40% of individuals with type II diabetes and a staggering 86% of those with type I diabetes exhibit signs of DR. Additionally, rural areas in China report DR prevalence rates as high as 43%.[3] Major ophthalmological organizations, such as the American Academy of Ophthalmology and the International Council of Ophthalmology, strongly recommend regular eye examinations for the early detection of DR.[4]

Vision loss associated with DR progresses gradually as the disease advances, typically categorized into two primary stages: non-proliferative diabetic retinopathy (NPDR) and proliferative diabetic retinopathy (PDR). PDR is marked by neovascularization or hemorrhages within the vitreous or pre-retinal regions.[4] The American Academy of Ophthalmology further classifies DR into five stages based on severity. The initial stage, known as "no apparent retinopathy," indicates a normal retinal examination with no visible signs of the disease. The subsequent stages—mild, moderate, and severe NPDR—represent progressively worsening retinal damage. Mild NPDR is characterized by the presence of microaneurysms without significant additional damage. As the condition advances to severe NPDR, it is identified by 20 or more hemorrhages per retinal quadrant, accompanied by intraretinal microvascular abnormalities and venous beading in at least two quadrants. Moderate NPDR falls between the mild and severe stages, indicating a moderate level of retinal damage. The final stage, PDR, is the most advanced form of diabetic retinopathy and involves the growth of new, abnormal blood vessels on the retina, known as retinal neovascularization, or bleeding into the vitreous humor (vitreous hemorrhage), or between the retina and the vitreous membrane (pre-retinal hemorrhage).[5] To assess the severity of DR, retina specialists often employ a diagnostic procedure called fluorescein angiography, which involves the injection of a fluorescent dye to visualize the blood vessels in the retina.[6][7]

The early detection and treatment of DR are crucial in preventing vision loss and maintaining eye health. As the prevalence of diabetes continues to rise, the importance of regular eye examinations and timely intervention cannot be overstated. Common symptoms of DR include blurry vision, the appearance of floaters or flashes, and vision loss. The condition primarily affects the macula due to metabolic changes in the retinal blood vessels. These changes lead to abnormal blood flow and leakage of blood components onto the retina. This causes the retinal tissues to swell, consequently causing clouded or blurred vision. Typically, DR impacts both eyes and, if left untreated, can progress to diabetic maculopathy. This advanced form of the disease leads to significant vision impairment and potentially complete blindness if diabetes is not properly managed over an extended period.[9]

The progressive nature of DR means that if left untreated can cause vision deterioration, ultimately leading to irreversible blindness. Over time, the disease results in the formation of retinal lesions due to ruptured retinal blood vessels. These lesions include microaneurysms (MAs), hemorrhages (HEs), exudates (EXs), and cotton wool spots (CWSs). Additionally, other abnormalities such as fibrotic bands, intra-retinal microvascular abnormalities (IRMAs), neovascularization on the disc (NVD), neovascularization elsewhere (NVE), and tractional bands also develop.[10][11] These lesions severely affect the fundus, which is the back of the eye, and their timely detection is essential for identifying the various stages of DR. [12]

DR's impact is not limited to vision alone. It can significantly affect the quality of life, limiting daily activities and increasing the risk of depression and anxiety. Hence, managing blood sugar levels effectively, coupled with regular eye exams, is crucial for those with diabetes to prevent the onset and progression of DR. Fluorescein angiography, a diagnostic procedure used by retina specialists, involves injecting a fluorescent dye into the bloodstream to visualize the blood vessels in the retina.[13] This method is instrumental in assessing the extent and severity of DR and guiding appropriate treatment plans.

Early detection of Diabetic Retinopathy (DR) is crucial for preventing severe vision loss and involves the meticulous monitoring of key retinal structures. These structures include the optic disc (OD), retinal blood vessels (RBVs), fovea, and macula. The initial steps in this process often involve pupil dilation using medically approved contrast agents, which are injected into the retina. This facilitates the capturing of detailed images of the fundus through techniques such as fluorescein angiography (FA) and mydriatic fundus photography.[13] These images are then carefully analyzed to detect early signs of DR. Traditional diagnostic methods[14] for DR include a variety of sophisticated imaging and analysis techniques. These methods encompass biomicroscopy, fundus imaging, retinal thickness analysis, scanning laser ophthalmoscopy (SLO), adaptive optics, retinal oximetry, optical coherence tomography (OCT), OCT angiography, and Doppler OCT. While these approaches provide comprehensive insights into retinal health and the presence of DR, they are often labor-intensive and time-consuming. Additionally, they require the expertise of highly trained specialists to interpret the results accurately. This reliance on specialized personnel is particularly challenging in regions like India, where the ratio of ophthalmologists to patients is approximately 1:10,000.[15] The practical challenges associated with traditional diagnostic methods underscore the urgent need for automated, intelligent detection systems. Such systems should be capable of conducting early screenings for DR efficiently and accurately. Leveraging large datasets of fundus images, these advanced systems can utilize machine learning and artificial intelligence to analyze retinal images, identify early signs of DR, and provide timely diagnoses. This approach has the potential to significantly improve early diagnosis rates and, consequently, preserve the vision of diabetic patients through timely medical intervention. Integration of automated detection systems into routine healthcare could reduce the burden on ophthalmologists, allowing them to focus on more complex cases. The use of AI-driven technology in DR screening can also ensure a broader reach, making high-quality eye care accessible to a larger population, especially in underserved areas.

The evolution of machine learning has led to an increase in the number of computer-assisted systems for detecting diabetic retinopathy (DR). These systems utilize algorithms like Support Vector Machines (SVMs), Neural Networks, and Decision Trees. However, traditional machine learning algorithms often struggle when faced with the complexity of large-scale, high-dimensional data such as medical images. They demand significant computational resources and lack the necessary flexibility and domain-specific insights to accurately represent data. The evolution of deep learning has been a game-changer offering an advanced approach to handling complex tasks. Deep learning excels at extracting deep features and processing high-dimensional data. Its frameworks are highly scalable and capable of learning domain-specific knowledge, enabling them to make reliable decisions that often surpass the performance of traditional models.

Various techniques like Convolutional Neural Networks (CNN)[16][17], Deep Convolutional Neural Networks (DCNN), and architectures such as AlexNet, VGG16, VGG-19[18], GoogLeNet, ResNet, DenseNet, and Inception have been utilized for deep feature extraction and image classification. Additionally, methods including Autoencoders, Restricted Boltzmann Machines (RBMs), networks, and Deep Reinforcement Learning (DRL) have also been explored.[19]

Despite these advancements, challenges persist, particularly due to the lack of high-quality annotated datasets, variations in image quality, and class imbalances that complicate model development.

This paper presents a comparison of U-Net and Multi-stacked U-Net architectures across various datasets, to analyze and find a saturation point for the number of stacks in the U-Net architecture for which the model performs the best. We have used 3 distinct and reliable datasets, APTOS-2019, IDRiD, and Diabetic Retinopathy Detection by Sachin Kumar, all available on Kaggle.

The Asia Pacific Tele-Ophthalmology Society (APTOS)[35] dataset comprises 3,662 retinal images collected from participants in rural India, with data collected by Aravind Eye Hospital in India, focusing on subjects from rural areas where medical screening is challenging. The images are categorized into cases of both non-proliferative diabetic retinopathy (NPDR) and proliferative diabetic retinopathy (PDR). The images vary in resolution, with common dimensions including 2416x1736, 1050x1050, and 819x614 pixels in RGB format.

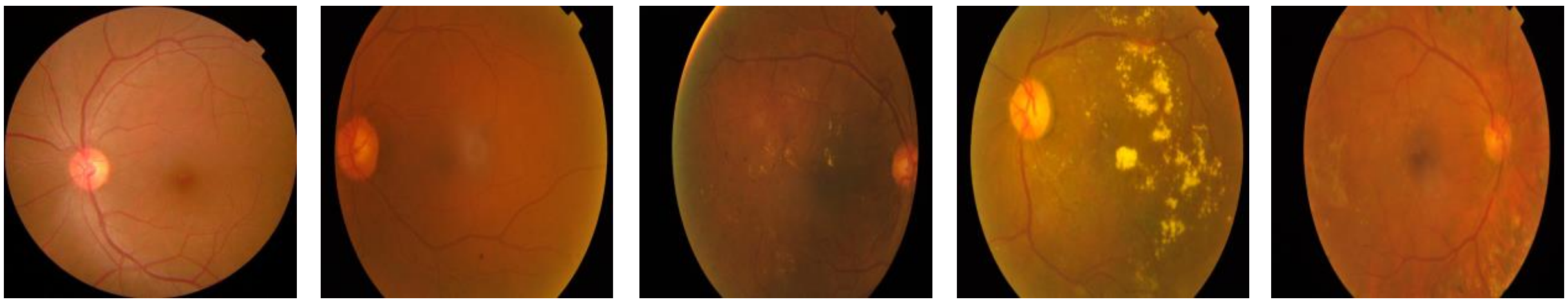

*Figure 1.1* *Figure 1.2* *Figure 1.3* *Figure 1.4* *Figure 1.5*

*Figure 1: APTOS Raw Data: 1.1 No Diabetic Retinopathy (No DR), 1.2 Mild Diabetic Retinopathy (Mild DR), 1.3 Moderate Diabetic Retinopathy (Moderate DR), 1.4 Severe Diabetic Retinopathy (Severe DR), and 1.5 Proliferative Diabetic Retinopathy (Proliferative DR).*

The Indian Diabetic Retinopathy Image Dataset (IDRiD)[36] comprises of images created from real clinical exams at an eye clinic in Nanded, Maharashtra, India, by a retinal specialist using a Kowa VX-10 alpha digital fundus camera. These images have a 50º field of view, a resolution of $4288 \times 2848$ pixels, and are stored in jpg format. The dataset includes 516 images, categorized into five Diabetic Retinopathy (DR) classes and three Diabetic Macular Edema (DME) classes, following international clinical standards. The dataset includes pixel-level annotations for various DR lesions such as microaneurysms, hemorrhages, hard exudates, and soft exudates, as well as optic disc annotations. Each image is graded for disease severity based on the International Clinical Diabetic Retinopathy Scale.

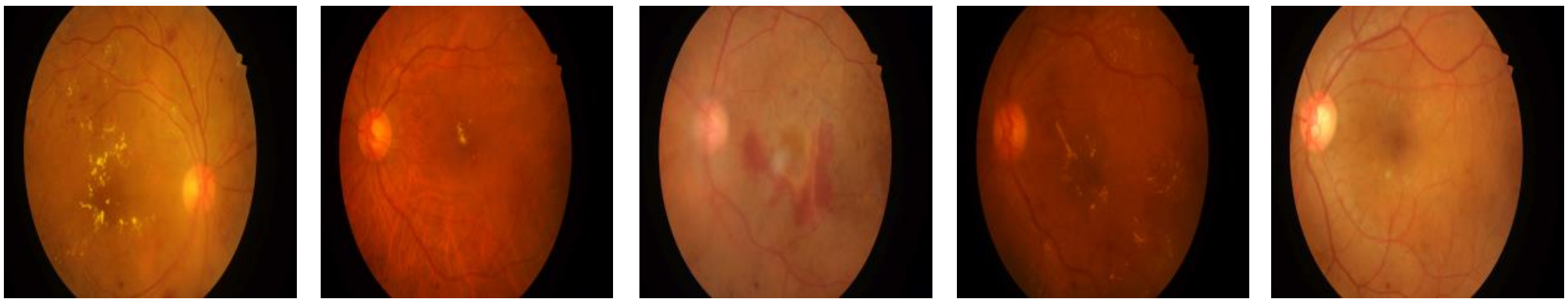

*Figure 2.1* *Figure 2.2* *Figure 2.3* *Figure 2.4* *Figure 2.5*

*Figure 2: IDRiD Raw Data: 2.1 No Diabetic Retinopathy (No DR), 2.2 Mild Diabetic Retinopathy (Mild DR), 2.3 Moderate Diabetic Retinopathy (Moderate DR), 2.4 Severe Diabetic Retinopathy (Severe DR), and 2.5 Proliferative Diabetic Retinopathy (Proliferative DR).*

The Sachin Kumar Diabetic Retinopathy dataset[37] consists of 2750 retinal images that were typically collected from clinical settings and are used to identify signs of diabetic retinopathy in retinal fundus images. The images are resized to 224x224 pixels for consistency and ease of processing and are all stored in either jpg or png format. This dataset includes annotations for various lesions such as microaneurysms, hemorrhages, hard exudates, and soft exudates, as well as optic disc annotations.

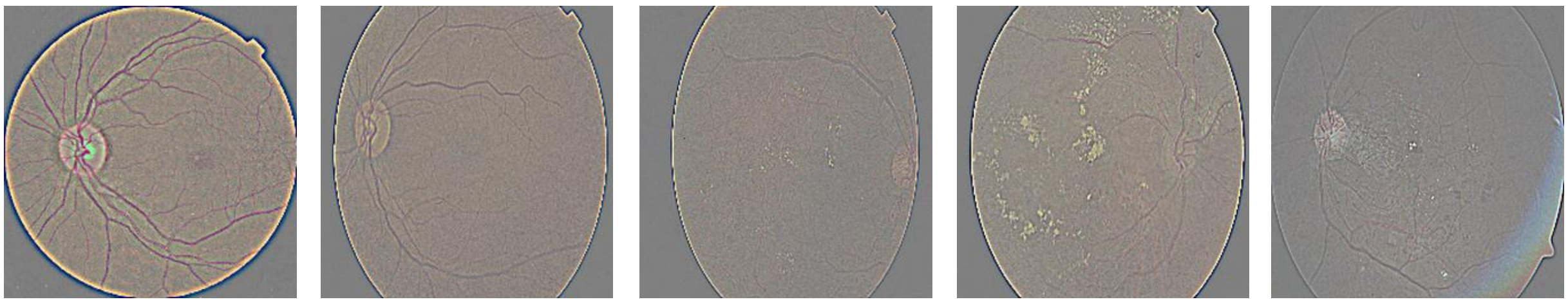

*Figure 3.1* *Figure 3.2* *Figure 3.3* *Figure 3.4* *Figure 3.5*

*Figure 3: Sachin Kumar Raw Data: 3.1 No Diabetic Retinopathy (No DR), 3.2 Mild Diabetic Retinopathy (Mild DR), 3.3 Moderate Diabetic Retinopathy (Moderate DR), 3.4 Severe Diabetic Retinopathy (Severe DR), and 3.5 Proliferative Diabetic Retinopathy (Proliferative DR).*

The original images across various datasets were initially in RGB format. For preprocessing, these images were resized to 256x256 pixels and converted to grayscale, resulting in dimensions of 256x256x1. This conversion helps simplify the data while retaining essential features necessary for model training. To augment the dataset, horizontal flips of the images were generated. Each flipped image retained the same classification label as its original counterpart. This augmentation technique effectively doubled the dataset size, providing the model with more diverse training examples.

UNET was chosen for its capability to use convolutional layers and max pooling operations to extract distinct features and patterns from the retinal images. The input layer of the Convolutional Neural Network (CNN) was configured to accept images of 256x256x1 dimensions. In a CNN, the convolutional layers are composed of multiple filters that are convolved with the input images to extract various features. These filters help the network learn to identify different patterns and structures within the image data. Following the convolutional operations, max-pooling layers were employed. Max pooling involves a smaller matrix, or kernel, moving across the convolutional layer to down-sample the input representation, reducing its dimensionality while preserving important features.

By repeatedly applying 2D convolution and max pooling layers, the network expands the image dimensions through learned feature hierarchies. Each convolution operation captures finer details, while max pooling reduces spatial dimensions, making the computation more efficient. The architecture can vary in the number of convolutional and pooling layers, depending on the complexity and requirements of the task. Ultimately, these layers feed into dense layers, also known as fully connected layers, where each neuron is connected to every neuron in the previous layer. This densely connected structure is crucial as it enables the network to combine all the learned features and make accurate predictions.

The design of the network, including the number and arrangement of layers, significantly influences the final output size and shape. The dense layers aggregate information from the entire image, allowing the model to make informed decisions based on all the features extracted during the convolution and pooling stages. This meticulous design and preprocessing approach ensure that the CNN can effectively learn from the dataset and generalize well to new, unseen data, enhancing its performance in detecting and classifying diabetic retinopathy.

## II. LITERATURE SURVEY

In the research conducted by Khan Z et al., a modified architecture of VGG-16 was implemented for the detection of Diabetic Retinopathy in its five categories. Here, the EyePacs dataset from Kaggle was used to train the model. It consists of 88k images, out of which 35K are classified into five classes starting from No DR to Proliferate DR and the rest of them are unclassified. The model architecture involved a VGG-16, a spatial pooling layer (SPP) and a Network in Network (NiN) to achieve a highly nonlinear scale-invariant deep model named VGG-NiN. The NiN layer is build using components called mlpconv which are stacked together to achieve the non linearity in the model. The mlpconv works similar to multilayer perceptron in the terms of backtracking and weight modification. The proposed model was capable of processing images of any scale due to the properties of the SPP Layer. Further the stacking of the NiN layers adds a component of non-linearity to the model training making it better for classification. The hyper parameters chosen for the training of this model involve , Batch size of 8, initial learning rate of 0.01 ,momentum of 0.9 and minimum learning rate of 0.0001. Finally the model was trained for 50 epochs to achieve an accuracy of 85% . Further the model achieved a Recall of 55.6%, Precision of 67%, Specificity of 91%, and an F1 Score of 60%.[20]

DenseNet is a CNN architecture in which each layer connects to every other layer, allowing for efficient feature reuse. This architecture mitigates the vanishing gradient problem by using an efficient and reduced set of parameters. DenseNet's design promotes information flow between layers, making it highly effective for image recognition tasks with fewer parameters compared to other traditional CNN architectures. In the paper by Mushtaq, G. et al., the architecture of DenseNet-169 is implemented on the APTOS (Asia Pacific Tele-Ophthalmology Society) dataset. The dataset consists of 5593 images with a heavy class imbalance. The class imbalance was corrected using the techniques of mirroring and rotation. A total dataset of 7000 images was generated for training. For the last layer of our model, a Global Average Pooling is added, followed by a dropout (0.5) and a five-node output layer to classify each class. The model uses Adam, with a sequential approach, to add and customize convolutional, dropout, dense, and optimization layers to prevent overfitting. The proposed model achieved an accuracy of 90% and a Cohen kappa score of 80%. The model also performed the same metrics in the Diabetic Retinopathy Detection dataset (2015).[21]
Tymchenko, B. et al. implemented the Albumenations library's set, optical distortion, grid distortion, piecewise affine transform, horizontal flip, vertical flip, random rotation, random shift, random scale, a shift of RGB values, random brightness and contrast, additive Gaussian noise, blur, sharpening, embossing, random gamma, and cutout to moderate the imbalance in the APTOS dataset. After image augmentation, a dataset of 13000 images was created for the model training. Further, a combination of the Messidor and IDRID datasets was also used for the network training. The network architecture proposed was a multi-task CNN model with an ImageNet-Pretrained Encoder and three decoders: classification, regression, and ordinal regression. Each of these decoders achieves a different task: the classifier detects the presence of the disease, the regression predicts the stage of the disease, and the ordinal rank regression predicts the severity of the disease. Each of these decoders is called a head, and they use specialized loss functions and a cosine-annealing schedule, with 5-fold cross-validation and an Adam optimizer. As a final part of the architecture, a linear regression model is fit to the outputs of these three heads, with weights of these three outputs being set to 0.33 and trained for five epochs to reduce MSE. With an ensemble of 3 CNN architectures (EfficientNet-B4, EfficientNet-B5, SE-ResNeXt50) and transfer learning as their final solution, they achieved metrics of QWK of 0.969, a macro F1 score of 0.840, an accuracy of 0.919, a sensitivity of 0.840, and a specificity of 0.981.[22]

It is not always possible for every other individual to have access or the funds to get access to a fundus camera to be able to use the said images for the model prediction. Hence, the need for simpler methods for image photography was demanded. Thus, since 2010, several companies have come up with pieces of equipment to attach to your smartphone to be able to capture fundus images using your smartphone. However, the drawback of such a method was the quality and the region of the eye covered in this image. The clarity of the vessels had deteriorated, and the eyeball was not fully captured; this led to further requirements in the automation of such a methodology with the help of ML. Thus, in the paper by Hacisoftaoglu RE et al., various models were implemented and tested for their performance for lower-quality images. This study first worked with renowned models like AlexNet, GoogLeNet, and ResNet50 architectures, using the transfer learning approach. Then, the images from several well-accepted datasets like EyePACS, Messidor, IDRiD, and Messidor-2 were cropped and blurred to imitate the image quality in smartphone-based images. Finally, the ResNet50 architecture is chosen, and the thus created dataset, along with smartphone synthetic data, is used to train the said model. This model achieved an accuracy of 98.6, 98.2% sensitivity, and 99.1% specificity, while its AUC was 0.9978 on an independent test dataset.[22]

The ADL (active deep learning)-CNN proposed by Qureshi I et al. is a seven-layer CNN model that follows an architecture similar to LeNet [34]. It has two convolutional and a max-pooling layer, followed by fully connected layers and a final Softmax layer classifying the image into the said five categories. The hyperparameters for the ADL-CNN model include a learning rate of 0.001, a learning decay ratio of 0.1, a weight decay of 0.0001, Leaky ReLU activation for layers 1, 3, 5, and 6, and Gaussian distribution for bias initialisation in the Softmax layer, with weight initialisation from predefined values. The SGD (Stochastic Gradient Descent) with a momentum value of 0.9 in the Nesterov mode is used for the updating of the weights. The architecture is trained on the EyePacs dataset. The input size is taken as 48 x 48 x 3, i.e., an RGB image of size 48 x 48. Preprocessing techniques like illumination correction, cropping and resizing, and colour normalisation are followed for the further normalisation of the dataset. The proposed model was compared with its performance with VGG-16, which it outperformed with an accuracy of 98%. in multiclass classification, average SE of 92.20%, SP of 95.10%, and F-measure of 93%. [23]
The model presented by A Ayala et al., is an improvised version of DenseNET121[33] for the multiclass classification of DR. The architecture implemented three blocks of the layers Conv2D, Dense, Conv2D and Max Pooling. Then this is connected to 2 layers of Fully connected layers are implemented for Classification. It Publicly available datasets of Messidor-2 and APTOS were used for the model training. The architecture has a detection algorithm that uses bounding boxes to locate and classify the lesions in retinal images. This improves the model's capability to identify the various stages of DR. The model trained on APTOS achieved 81% validation accuracy, 59% testing accuracy on Messidor, and outperformed Messidor with higher precision, recall, and F1-score across most classes, except for Severe NPDR and PDR. The model outperformed in the APTOS dataset with a Precision of 0.96, Recall of 0.99, and F1-Score of 0.97. [24]

Anas Bilal, Guangmin Sun, Sarah Mazhar, Azhar Imran, and Jahanzaib Latif propose a two-stage framework—Stage One utilizes two distinct U-Net models for optic disc (OD) and blood vessel (BV) segmentation to capture essential retinal features aiding in DR detection; Stage Two classifies DR severity levels using a transfer learning-based VGG-Net model and utilizes the segmented images from the first stage to improve classification accuracy on these important retinal biomarkers. This architecture aims to resolve the challenges in the classification of diabetic retinopathy in such a manner that the study segregates informative regions in the retina that are significant for determining different degrees of severity in diabetic retinopathy, by a higher accuracy as a result of a better diagnosis. The collection of datasets, namely EyePACS-1, Messidor-2, and DIARETDB0 are used, providing a full collection of 11, 841 color fundus images in RGB format. All images are resized uniformly to 256 × 256 pixels so that neural networks can process the images uniformly. This ensures that data quality is consistently attained and that learning in the neural network from identical inputs is optimized so that the model performs well on all datasets. The model obtained an average accuracy score of 96.60% on EyePACS-1; 93.95% on Messidor-2; and 92.25% on DIARETDB0. [26]

In the paper written by, Dinial Utami Nurul Qomariah, Handayani Tjandrasa, and Chastine Fatichah a novel architecture called MResUNet was implemented, which is an improvised version of the existing UNet architecture focused on retinal microaneurysm segmentation improvement. To overcome the pixel imbalance issue between the microaneurysm and its background pixels, this new structure includes residual units with identity mapping which enables deeper learning and retention of pixels that are important features of the structure. To address this challenge further, the authors adopt an adapted weighted cross entropy loss function, which efficiently ranks the microaneurysms for the most loss caused by the background while ensuring microaneurysms are prioritized. For evaluating the model, two retinal datasets are investigated in this study. The first part of the study utilizes the IDRiD dataset which contains 81 RGB format images captured using a digital fundus camera with a resolution size of 4288 x 2848 pixels. These images are enhanced using are cumulative histogram equalization and contrast limited adaptive histogram equalization (CLAHE) to achieve a higher contrast in the images for easier detection. The second part of the study utilizes the DiaretDB1 dataset, which contains 89 color images also of patients with diabetic retinopathy, providing the study with a wide range of the condition. MResUNet managed to reach a sensitivity score of 61.96% when analyzing the first IDRiD dataset and 85.87% when evaluating the DiaretDB1 dataset. The proposed model also achieved an accuracy of 99.75 for DiaretDB1 dataset and an accuracy of 99.76 for the IDRiD dataset.[27]

The study proposes a hybrid deep-learning approach for Diabetic Retinopathy (DR) detection from fundus images, authored by Muhammad Mohsin Butt, D. N. F. Awang-Iskandar, Sherif E. Abdelhamid, Ghazanfar Latif, and Runa Alghazo. The architecture invokes transfer learning with the aid of the pre-trained Convolutional Neural Networks (CNNs)-Well, GoogleNet[30], and ResNet-18[31]. The last layer (Fully connected Layer) of these models that are used for classification are ommited and the features generated are extracted. 1000 features are extracted from each of the networks. They are merged to yield a hybrid feature vector map, with a total of 2000 features, which is then fed into classifiers like Support Vector Machine (SVM), Random Forest (RF), Radial Basis Function (RBF), and Naïve Bayes (NB). This hybrid model supports both binary and multiclass classification of diabetic retinopathy; the APTOS dataset is used for the given models training. The images are resized to 224 * 224 * 3 that matches the input size of the predefined Architectures of ResNet and GoogLeNet. The SVM based classifier achieved the highest overall performance, with an Accuracy of 97.39%, Precision and Recall of 97.40%, and an F-Measure of 97.40%. The RBF based classifier closely followed, with an Accuracy of 96.98%, Precision and Recall of 97.00%, and an F-Measure of 97.00%. [25]

The Research conducted by Tiwalade Modupe Usman, Yakub Kayode Saheed, Djitog Ignace, and Augustine Nsang presents a Deep Learning Multi-Label Feature Extraction and Classification (ML-FEC) model for diabetic retinopathy (DR) lesion detection and classification in color fundus photographs (CFPs). Transfer learning is applied by utilizing three state-of-the-art Convolutional Neural Network (CNN) architectures: ResNet50 [31], ResNet152 [31], and SqueezeNet1 [32]. The pre-trained models are fine-tuned, where only their last layers are exponentially reconfigured to be more adaptable to the distinctive properties of the DR dataset. PCA is employed by the remaining part of this model for extraction, i.e., multi-label, which aids in simultaneously detecting multiple lesions in one image. The dataset used for detecting diabetic retinopathy (DR) and clinically significant macular edema (CSME) contains 1445 high-quality fundus images acquired from the Department of Ophthalmology at Rawalpindi Medical University, Pakistan. These 2976 × 2976-pixel images were annotated by a panel of five senior ophthalmologists, detailing six DR stages and various retinal lesions, including exudates, hemorrhages, and neovascularization. The six stages of DR include No DR, Mild NPDR, Moderate NPDR, Severe NPDR, Early PDR, and high-risk PDR, as well as lesions inclusive of exudates and hemorrhages. The experiments showed that the ResNet152 architecture had the highest accuracy-minimum model, being 94.4 percent-sensitivity, which was 76.35 percent, F1 score of 72.87 percent, and Hamming loss of 0.0560. [28]

Convolutional Neural Networks (CNNs), as discussed in the aforementioned papers, have proved to have a remarkable ability to identify feature maps in images, making them effective for a wide range of image-based tasks, including classification and detection. However, CNN architectures face limitations; for example, the pooling layer can lead to information loss, and substantial data is required for training. Capsule Networks were developed as an alternative to address these issues, preserving spatial hierarchies in features and improving robustness to input variations. Therefore, G. Kalyani et al. provide a novel architecture of Capsule Networks (CapsNet) for diabetic retinopathy (DR) detection and classification from fundus images. The CapsNet architecture comprises three major parts: a convolutional layer for feature extraction, a primary capsule layer for processing the extracted features, and a class capsule layer for classification by dynamic exercises. The convolutional layer extracts features from the input fundus images; the primary capsule layer reshapes and squashes these features into vectors; and the class capsule layer ultimately classifies the images into several grades of DR based on the output of the primary capsules. The outputs of the class capsule are softened by passing them through a softmax layer to produce probabilities across all class images. The Messidor dataset is used for the evaluation of the proposed model. The images are provided in TIFF format and are visualized at different resolutions, ranging from 1,440× 960 to 2,304× 1,536. Each image is also labelled by retinopathy grades ranging from 0 (healthy retina) to 3 (severe diabetic retinopathy), corresponding to the appearance of some particular lesions, such as microaneurysms or haemorrhages. It achieves an accuracy of 97.98% on healthy retina samples; for grade-1 diabetic retinopathy, it achieves an accuracy of 97.65%; for grade-2 type, it also achieves an accuracy of 97.65%, except for grade-3 type, where it gets the highest accuracy and reaches up to 98.64%. [29]

## III. PROPOSED METHODOLOGY

The U-Net architecture excels in pixel-level localization by using upsampling layers in the expansive path, paired with max-pooling in the contracting path, thus increasing its applicability in domains that require precise localization of features. While similar localization might be achieved using fully connected layers, this would result in a model that is computationally heavy, highly nonlinear, and generally less effective at preserving spatial relationships. Therefore, this design is well-suited for applications requiring detailed segmentation, as the skip connections preserve high-resolution spatial features throughout the network.

The U-Net architecture is composed of a contracting (downsampling) and an expanding (upsampling) path. In the downsampling path, each encoder block consists of two convolutional layers with ReLU activation, followed by a max pooling layer that reduces the spatial dimensions while increasing the number of feature channels. Each encoder block uses more filters as the network deepens, starting with 64 filters in the first layer, followed by 128 and 256 in subsequent layers. This progressive increase allows the model to learn more complex features at each level. Starting with an input of 256x256, the image size reduces through each encoder block—from 128x128 after the first max-pooling layer to 64x64, and finally to 32x32 in the last encoder block.

The output features from each encoder block are saved for skip connections in the upsampling path, where they are concatenated with corresponding layers to preserve spatial context. These skip connections enable better localization by merging low-level spatial information from earlier layers with high-level contextual information. Before entering the upsampling path, the features pass through two additional convolutional layers, which refine the features before reconstruction in the upsampling path.

In the upsampling path of the U-Net architecture, each block is a decoder block, responsible for gradually reconstructing the spatial resolution and refining feature maps. Each decoder block consists of a 2D transposed convolution layer, which increases the spatial resolution by learning upsampling transformations and restoring spatial details lost in the downsampling path. This transposed convolution layer is followed by a concatenate layer that combines features from the transposed convolution with corresponding features from the downsampling (encoder) path, using saved feature maps via skip connections. These skip connections preserve fine-grained spatial information from the encoder, enhancing the network's ability to localize and identify features accurately.

In each decoder block, convolution layers with ReLU activation further refine the upsampled features, with the number of filters set in reverse order to the encoder's sequence: 256, 128, and 64. This filtering pattern mirrors the feature extraction hierarchy of the encoder path while adapting it to reconstruct spatial details in the output image. The spatial size of each layer in the upsampling path reverses the downsampling path's progression, ultimately restoring the original image resolution, such as 256x256 pixels for a grayscale image.

At the bottleneck, a double set of convolutional layers with a high number of filters (1024 in this case) is used to capture the most complex features before entering the upsampling path. Finally, the network outputs a reconstructed image with the same spatial dimensions as the input. It is important to note that the image output at this point before it is sent into classification is often used for the purpose of image segmentation and recognizing the region of localization. For classification tasks, a flattened layer converts the image into a single tensor, which is then passed to three dense layers (sizes 256, 128, and 5) to produce a final prediction. This design makes U-Net suitable for pixel-wise classification, as it learns fine-grained localization alongside high-level context, ideal for image segmentation and multi-class tasks.

In an n-stack U-Net architecture, the 256x256 output of the upsampling section from each U-Net layer is sent as input to the downsampling part of the next U-Net stack. This process allows the model to refine features in stacks continuously, increasing its capability to capture and reprocess multi-scale and fine-grained spatial details. Iterative processing of the image helps the model to learn deeper and more complex hierarchical features across n levels and is thus thought to be nicely suited for high-resolution detailed tasks, e.g., segmentation and classification.

The final 256x256 output of the nth U-Net stack is then passed to the Flatten and Dense layers. The flattening layer converts the entire image into a 1D tensor, which then feeds into Dense layers for classification. Each Dense layer further refines the pattern recognition for the feature representation, with the last Dense layer outputting the desired number of classes, which in this case is five.

Each U-Net stack uses Conv2D (2D convolutional) layers, the essential building blocks for image analysis and other tasks in convolutional networks. Convolution operations are performed on the preprocessed set of images using kernels, which are a set of learnable filters. These filters slide over the input data, computing element-wise products, summing them up to produce feature maps that capture image patterns. Parameters for Conv2D include the number of filters, filter size (kernel size), padding, and activation function. Here, we use a filter size of 3x3 and padding set to 'same,' which maintains the spatial dimensions across layers

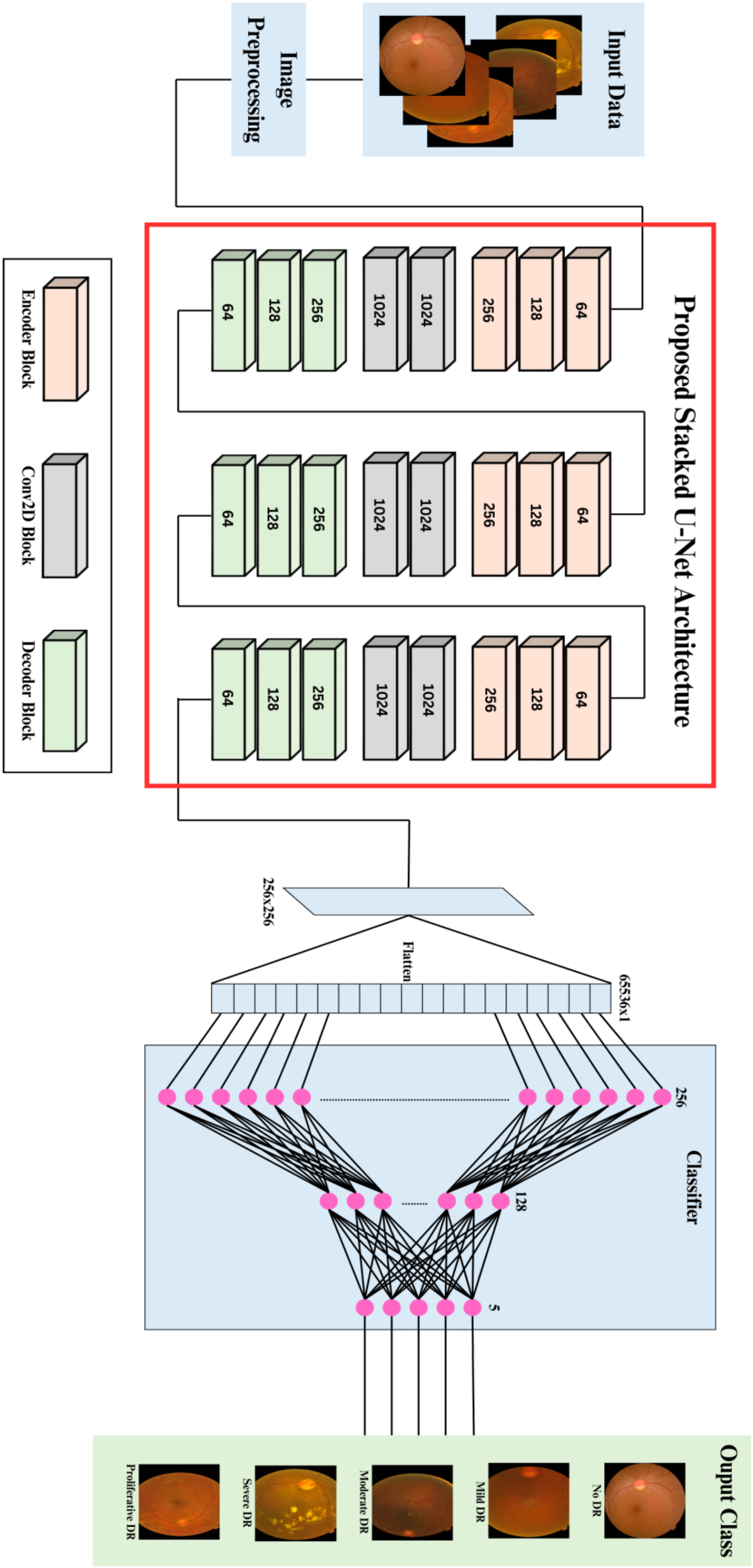

*Figure 4: Halo*

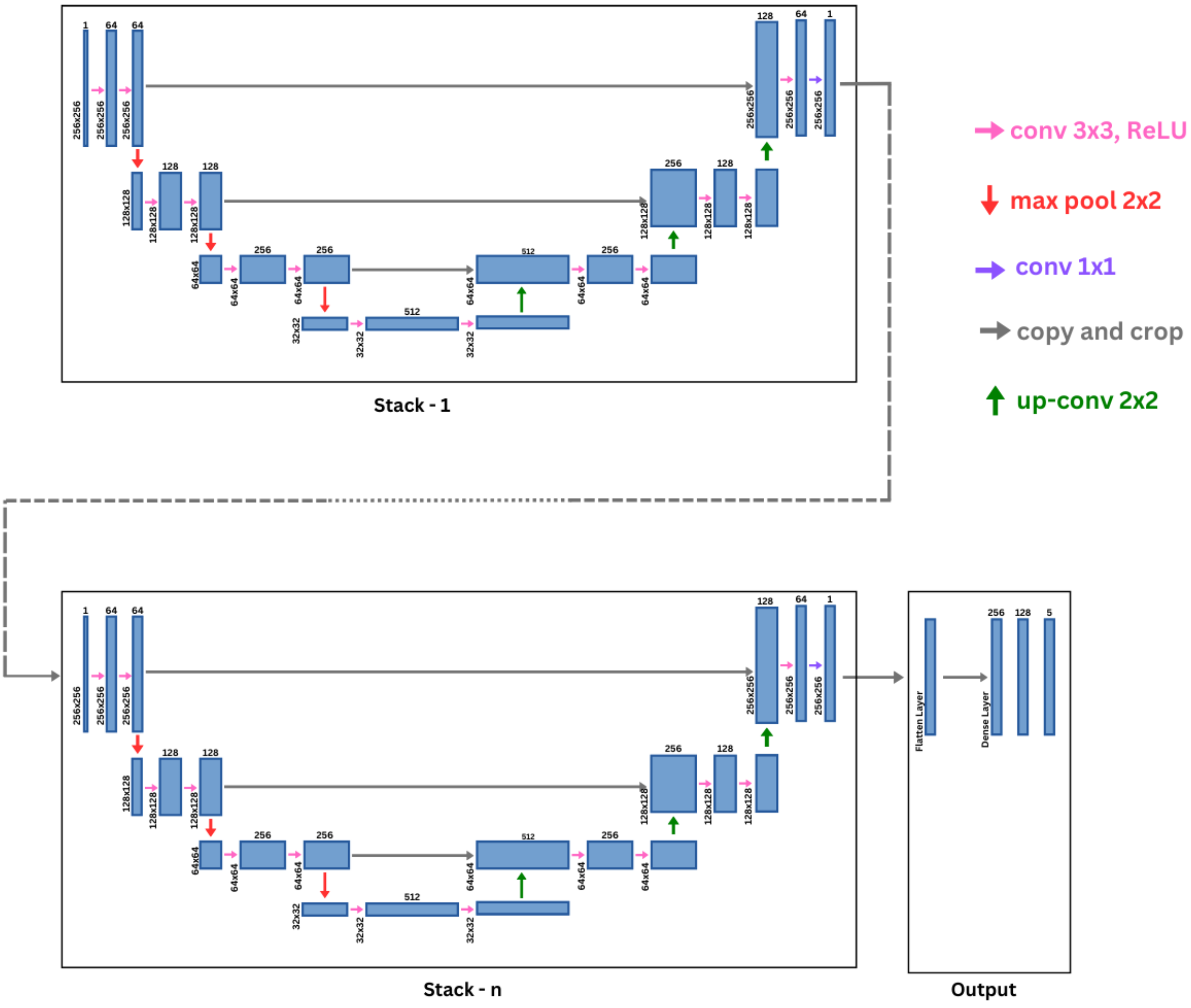

*Figure 5: n-Stack UNET Architecture*

Three open-source datasets were used for the evaluation of our proposed architectures, namely, APTOS, IDRiD, and a Kaggle-sourced diabetic retinopathy detection dataset by Sachin Kumar.

With respect to the APTOS dataset, augmentation by taking a vertical flip for all the images in the dataset. The images are all resized to 256 x 256 and converted to grayscale.

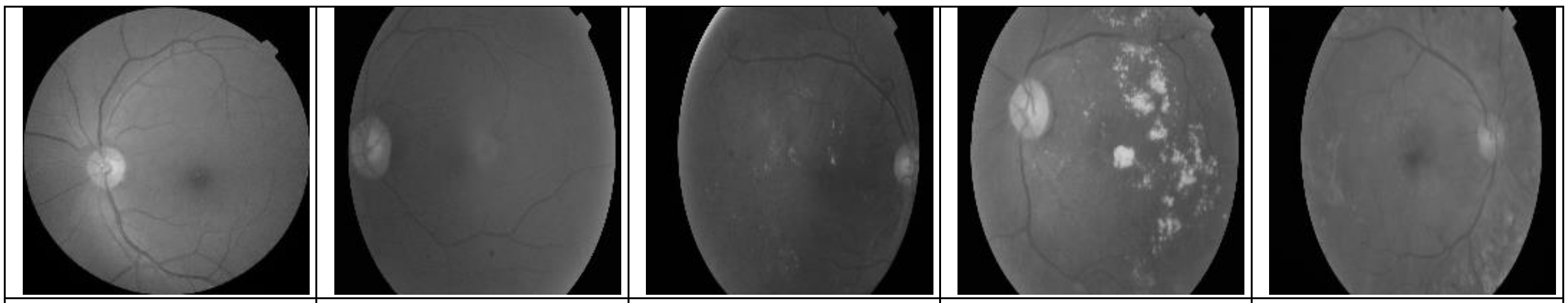

| *Figure 6.1* | *Figure 6.2* | *Figure 6.3* | *Figure 6.4* | *Figure 6.5* |
|---|---|---|---|---|

*Figure 6: APTOS Preprocessed: - Greyscale: 6.1 No Diabetic Retinopathy (No DR), 6.2 Mild Diabetic Retinopathy (Mild DR), 6.3 Moderate Diabetic Retinopathy (Moderate DR), 6.4 Severe Diabetic Retinopathy (Severe DR), and 6.5 Proliferative Diabetic Retinopathy (Proliferative DR).*

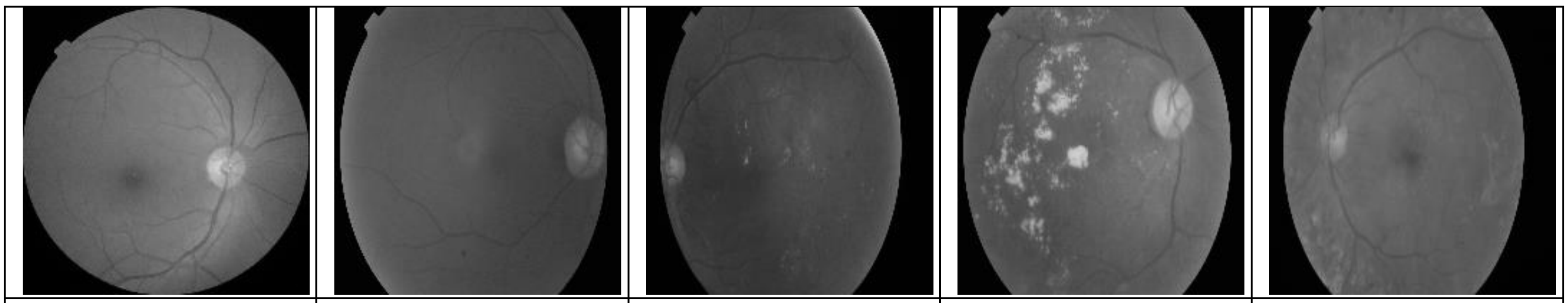

| *Figure 7.1* | *Figure 7.2* | *Figure 7.3* | *Figure 7.4* | *Figure 7.5* |
|---|---|---|---|---|

*Figure 7: APTOS Preprocessed - Vertical Flip: 7.1 No Diabetic Retinopathy (No DR), 7.2 Mild Diabetic Retinopathy (Mild DR), 7.3 Moderate Diabetic Retinopathy (Moderate DR), 7.4 Severe Diabetic Retinopathy (Severe DR), and 7.5 Proliferative Diabetic Retinopathy (Proliferative DR).*

In the case of IDRiD and Sachin Kumar Diabetic Retinopathy Datasets, augmentation techniques needed to be improvised considering that the dataset was much smaller in comparison with APTOS; thus, salt and pepper noise was added to the images with a strength of 0.04 along with the vertical flip of the image to triple the size of the original dataset. In the salt and pepper noise addition, the strength or amount of noise (ρ) is directly proportional to the Noise Density, which is decided based on Peak Signal is to Noise Ratio (PSNR).

$$PSNR(\rho) = const - 10log_{10}(\rho)$$

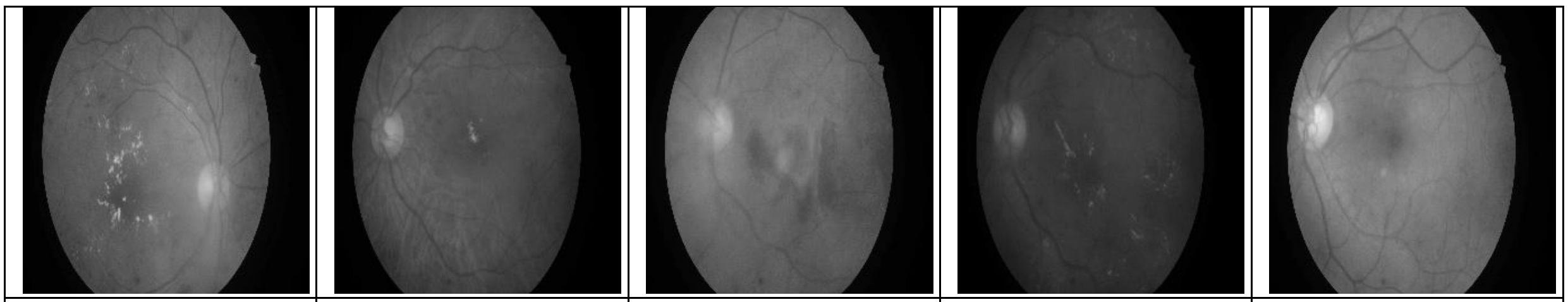

| *Figure 8.1* | *Figure 8.2* | *Figure 8.3* | *Figure 8.4* | *Figure 8.4* |
|---|---|---|---|---|

*Figure 8: IDRiD Preprocessed – Greyscale: 8.1 No Diabetic Retinopathy (No DR), 8.2 Mild Diabetic Retinopathy (Mild DR), 8.3 Moderate Diabetic Retinopathy (Moderate DR), 8.4 Severe Diabetic Retinopathy (Severe DR), and 8.5 Proliferative Diabetic Retinopathy (Proliferative DR).*

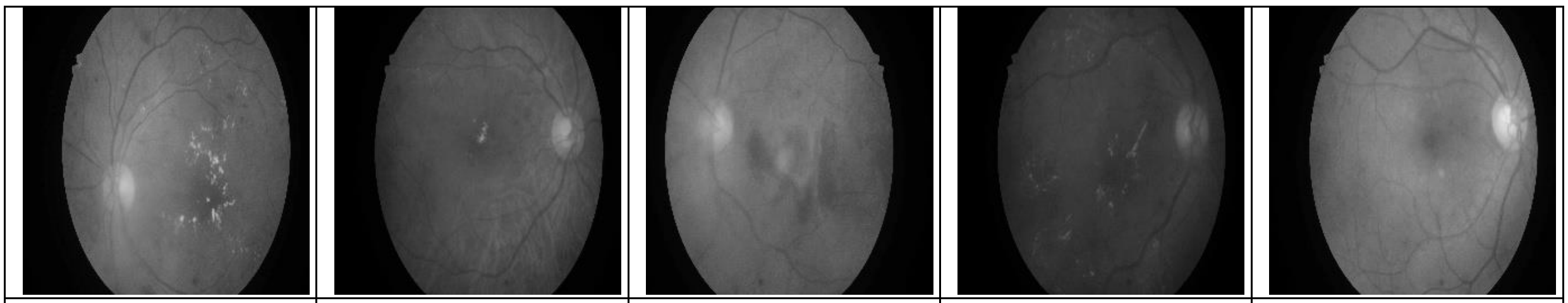

| Figure 9.1 | Figure 9.2 | Figure 9.3 | Figure 9.4 | Figure 9.5 |
|---|---|---|---|---|

*Figure 9: IDRiD Preprocessed – Vertical Flip: 9.1 No Diabetic Retinopathy (No DR), 9.2 Mild Diabetic Retinopathy (Mild DR), 9.3 Moderate Diabetic Retinopathy (Moderate DR), 9.4 Severe Diabetic Retinopathy (Severe DR), and 9.5 Proliferative Diabetic Retinopathy (Proliferative DR).*

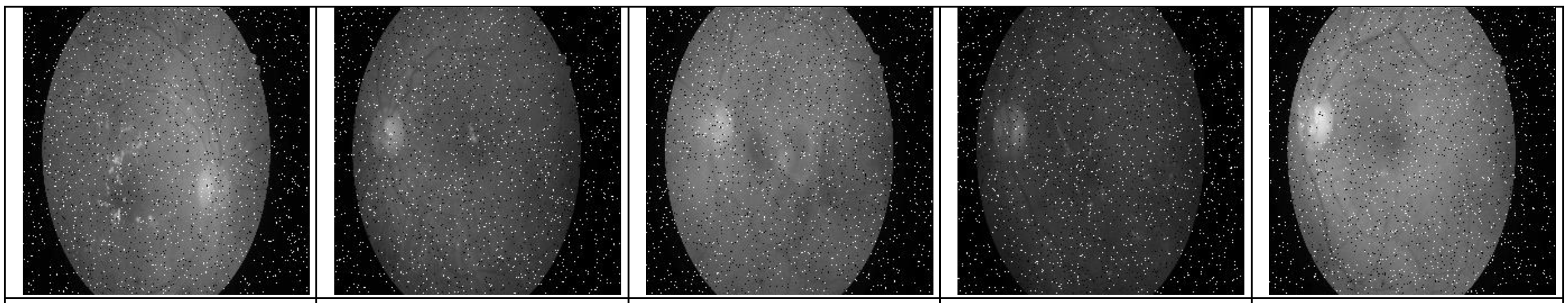

| Figure 10.1 | Figure 10.2 | Figure 10.3 | Figure 10.4 | Figure 10.5 |
|---|---|---|---|---|

*Figure 10: IDRiD Preprocessed – Salt and Pepper Noise: 10.1 No Diabetic Retinopathy (No DR), 10.2 Mild Diabetic Retinopathy (Mild DR), 10.3 Moderate Diabetic Retinopathy (Moderate DR), 10.4 Severe Diabetic Retinopathy (Severe DR), and 10.5 Proliferative Diabetic Retinopathy (Proliferative DR).*

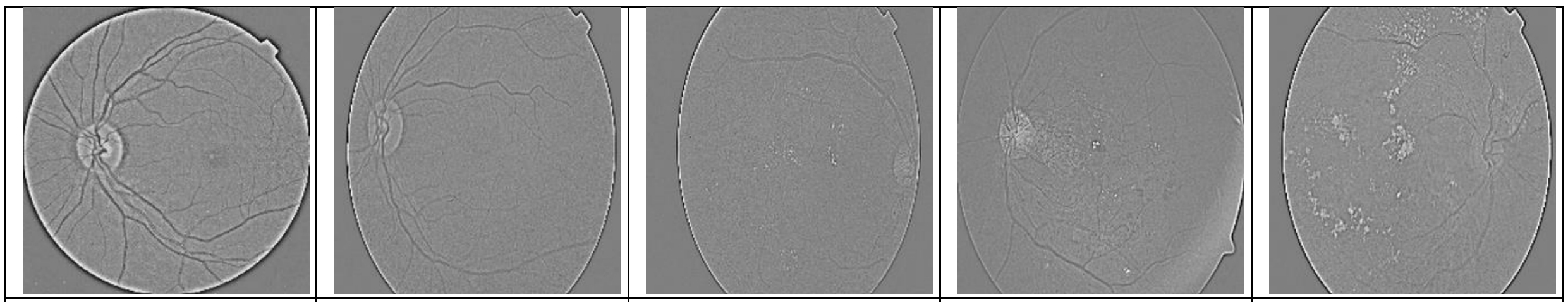

| Figure 11.1 | Figure 11.2 | Figure 11.3 | Figure 11.4 | Figure 11.5 |
|---|---|---|---|---|

*Figure 11: Sachin Kumar Preprocessed – Greyscale: 11.1 No Diabetic Retinopathy (No DR), 11.2 Mild Diabetic Retinopathy (Mild DR), 11.3 Moderate Diabetic Retinopathy (Moderate DR), 11.4 Severe Diabetic Retinopathy (Severe DR), and 11.5 Proliferative Diabetic Retinopathy (Proliferative DR).*

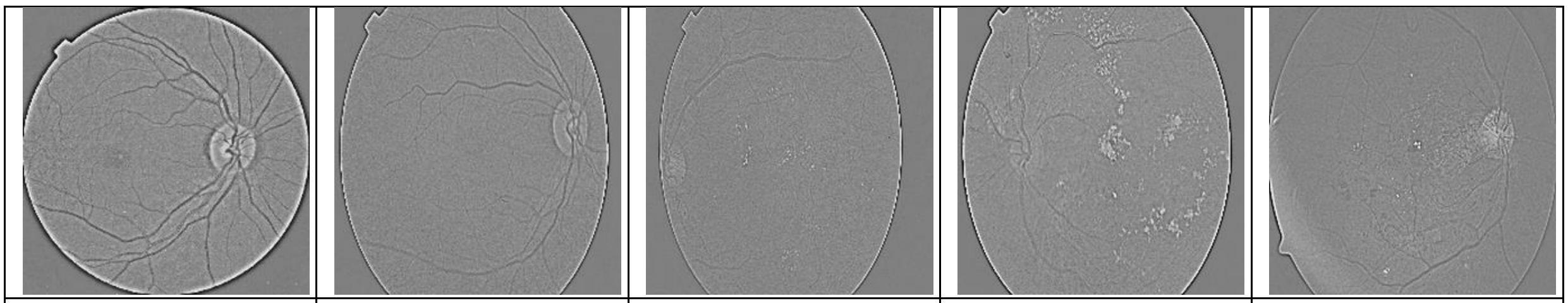

| Figure 12.1 | Figure 12.2 | Figure 12.3 | Figure 12.4 | Figure 12.5 |
|---|---|---|---|---|

*Figure 12: Sachin Kumar Preprocessed – Vertical Flip: 12.1 No Diabetic Retinopathy (No DR), 12.2 Mild Diabetic Retinopathy (Mild DR), 12.3 Moderate Diabetic Retinopathy (Moderate DR), 12.4 Severe Diabetic Retinopathy (Severe DR), and 12.5 Proliferative Diabetic Retinopathy (Proliferative DR).*

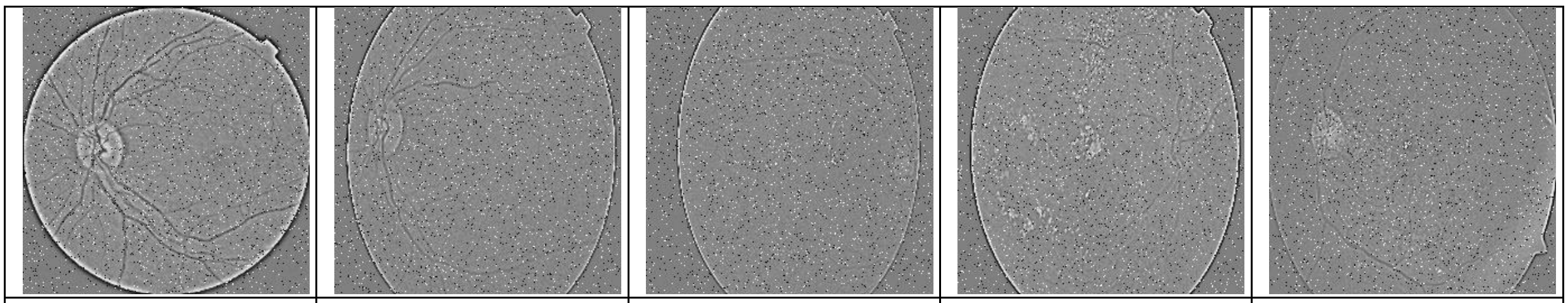

| Figure 13.1 | Figure 13.2 | Figure 13.3 | Figure 13.4 | Figure 13.5 |
|---|---|---|---|---|

*Figure 13: Sachin Kumar Preprocessed - Salt and Pepper Noise: 13.1 No Diabetic Retinopathy (No DR), 13.2 Mild Diabetic Retinopathy (Mild DR), 13.3 Moderate Diabetic Retinopathy (Moderate DR), 13.4 Severe Diabetic Retinopathy (Severe DR), and 13.5 Proliferative Diabetic Retinopathy (Proliferative DR).*

A train-test split of 0.2 was implemented across all datasets. With respect to optimizers, the Adam optimizer was used with a learning rate of $10^{-4}$ , an epsilon value of $10^{-6}$ and the amsgrad[40] convergence optimizer. A smaller value for learning rate helps us reach a more stable convergence point; however, it does run the risk of getting stuck at a local minima. However, increasing the learning rate leads to unstable convergence by overshooting the local minima. The epsilon value helps us control the learning rate of the model, whenever the gradients to be updated are way too small. Without this the metrics in the model often hit Not a Number (NaN) error while training a vastly varied dataset like the ones in this case. $10^{-6}$ is the most used value for the value of epsilon but higher values can be used based on requirement of stabilization. But one should be aware that a larger value of epsilon could potentially affect precision of the gradient updates. To understand the optimization offered by amsgrad one must understand the working of the moments in the training of a model, First moment estimate (m) is defined as the Exponentially weighted average of past gradients, While Second moment estimate (v) is defined as Exponentially weighted average of past squared gradients. By default, Adam updates the moment estimates (first and second moments) using a decaying average, but the amsgrad variant ensures that the second moment estimate (used for the adaptive learning rate) does not decrease over time. This can help prevent some theoretical issues in Adam where the learning rate can increase in some cases, leading to instability. However, using amsgrad any given scenario without its innate requirement would lead to making the training computationally expensive.

$$m_t = \beta_1 m_{t-1} + (1 - \beta_1) g_t$$

$$v_t = \beta_2 v_t + (1 - \beta_2) g_t^2$$

$$v_t = \max\left(v_{t-1}, v_t\right)$$

$$\theta_{t+1} = \theta_t - \frac{\eta}{v^{\frac{1}{2}} + \epsilon} m_t$$

## IV. METRICS

For multiclass classification problems, categorical cross-entropy[38] is the most efficient and suitable loss function to measure the difference between true class labels and predicted probabilities for each class.

$$Categorical\ Cross\ Entropy = -\sum_{i=1}^{i=N} y_{true_i} \cdot \log\left(y_{pred_i}\right)$$

$$Categorical\ Cross\ Entropy = -\sum_{i=1}^{i=N} y_i - \log(\overline{y_i})$$

The accuracy[39] calculates the proportion of times the model succeeded in making the correct predictions against total predictions. This metric shows the general performance, but it can be deceptive in the case of imbalanced datasets, because one could achieve high accuracy through a biased prediction favoring the majority class.

$$Accuracy\ =\ \frac{(TP\ +\ TN)}{(TP\ + TN\ + FP\ + FN)}$$

Precision[39] expresses the accuracy of positive predictions, showing how frequently a model has made positive predictions that are correct. So high precision is desired in situations where false positives incur a high cost, e.g. in medical diagnoses.

$$Precision\ =\ \frac{TP}{(TP\ + FP)}$$

Recall[39], on the other hand, computes the ability of the model to identify actual positive instances correctly. When there are high costs involved for false negatives, the need for high recall becomes evident. Recall expresses the capacity of a model to recognize almost all actual positive cases as such, providing highly cost-effective results in cases like fraud detection or disease diagnosis, where missing a positive case brings tremendous costs.

$$Recall\ =\ \frac{TP}{(TP\ + FN)}$$

The AUC (area under the ROC curve)[39] describes the power of a particular model to tell apart positive from negative classes at various thresholds for classification. AUC is based on the plot of true positive rate (recall) against the false positive rate. Higher values of AUC indicate better performance of the model. AUC is useful in model determination when performance assessment is required across various thresholds-in particular when doing so is a better evaluation than simple accuracy.

$$AUC\ =\ \int_0^1 TPR\ \cdot FPR\ \cdot dFPR$$

$$FPR\ = \frac{FP}{(FP\ + TN)}$$

$$FPR\ = \frac{FP}{(FP\ + TN)}$$

# V. RESULTS

The performance of our model was evaluated across three datasets: the APTOS-2019 Blindness Detection Dataset, the Indian Diabetic Retinopathy Dataset, and the Sachin Kumar Diabetic Retinopathy Dataset. For each dataset, the model's performance was assessed using various metrics including accuracy, Area Under the Curve (AUC), loss, precision, and recall across different stack configurations, and the results are tabulated below.

*Table 1: Metrics*

| Dataset | Stacks | Metrics | Accuracy | AUC | Loss | Precision | Recall |
|---|---|---|---|---|---|---|---|
| APTOS-2019 Blindness Detection Dataset | 1 | Training | 0.9914 | 1.0000 | 0.0168 | 0.9914 | 0.9910 |
| | | Validation | 0.9281 | 0.9859 | 0.3066 | 0.9338 | 0.9203 |
| | 2 | Training | 0.9883 | 0.9999 | 0.0187 | 0.9883 | 0.9883 |
| | | Validation | 0.9332 | 0.9880 | 0.2700 | 0.9380 | 0.9285 |
| | 3 | Training | 0.9842 | 0.9999 | 0.0297 | 0.9845 | 0.9838 |
| | | Validation | 0.9513 | 0.9905 | 0.1924 | 0.9578 | 0.9486 |
| | 4 | Training | 0.9911 | 0.9999 | 0.0221 | 0.9915 | 0.9911 |
| | | Validation | 0.9213 | 0.9816 | 0.3075 | 0.9249 | 0.9122 |
| Indian Diabetic Retinopathy Dataset | 1 | Training | 1.0000 | 1.0000 | 0.0011 | 1.0000 | 1.0000 |
| | | Validation | 0.9838 | 0.9999 | 0.0370 | 0.9838 | 0.9838 |
| | 2 | Training | 0.9960 | 1.0000 | 0.0057 | 0.9960 | 0.9960 |
| | | Validation | 0.9919 | 0.9998 | 0.0347 | 0.9919 | 0.9919 |
| | 3 | Training | 1.0000 | 1.0000 | 0.0072 | 1.0000 | 0.9960 |
| | | Validation | 0.9959 | 1.0000 | 0.0230 | 1.0000 | 0.9919 |
| | 4 | Training | 1.0000 | 0.9999 | 0.0178 | 0.9879 | 0.9879 |
| | | Validation | 0.9919 | 1.0000 | 0.0062 | 1.0000 | 1.0000 |
| | 5 | Training | 0.9980 | 1.0000 | 0.0065 | 0.9980 | 0.9980 |
| | | Validation | 0.9595 | 0.9996 | 0.0960 | 0.9595 | 0.9595 |
| Sachin Kumar Diabetic Retinopathy Dataset | 1 | Training | 0.9915 | 1.0000 | 0.0152 | 0.9915 | 0.9915 |
| | | Validation | 0.9783 | 0.9997 | 0.0586 | 0.9794 | 0.9771 |
| | 2 | Training | 0.9904 | 0.9999 | 0.0165 | 0.9907 | 0.9903 |
| | | Validation | 0.9798 | 0.9998 | 0.0489 | 0.9822 | 0.9786 |
| | 3 | Training | 0.9898 | 0.9999 | 0.0223 | 0.9898 | 0.9895 |
| | | Validation | 0.9823 | 0.9999 | 0.0565 | 0.9707 | 0.9821 |
| | 4 | Training | 0.9905 | 0.9999 | 0.0201 | 0.9906 | 0.9899 |
| | | Validation | 0.9819 | 0.9998 | 0.0455 | 0.9843 | 0.9807 |

In evaluating the detailed performance of our model across the three datasets, we observed notable results in the evaluation metrics.

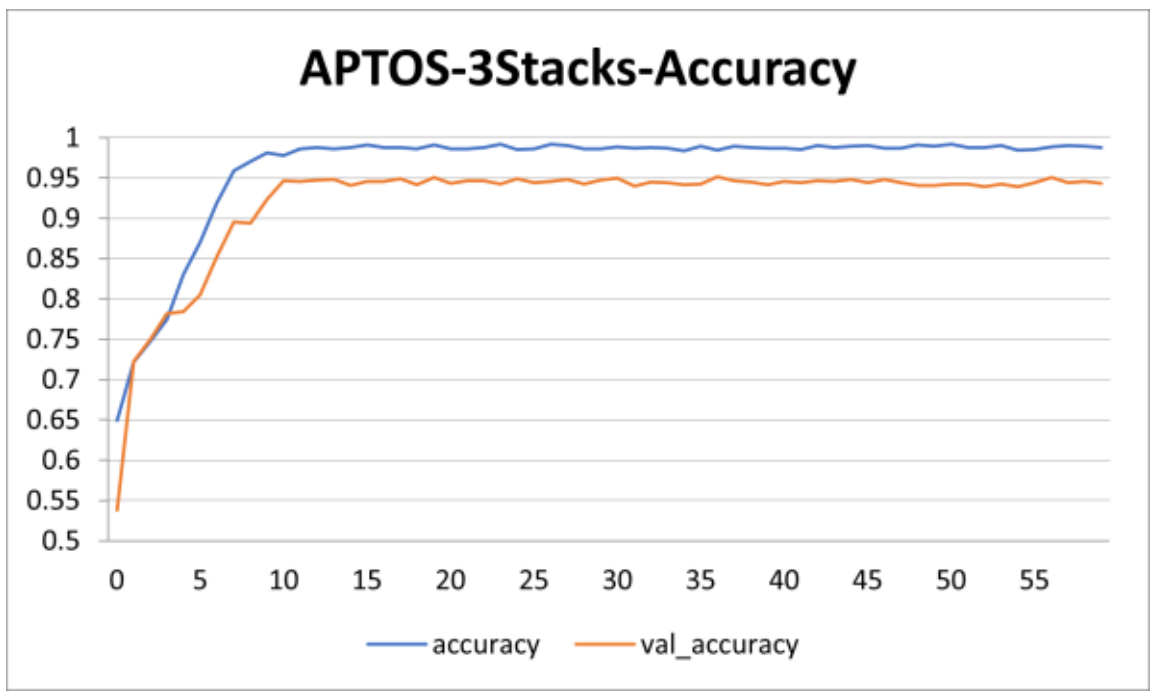

*Figure 14: APTOS-3Stacks-Accuracy*

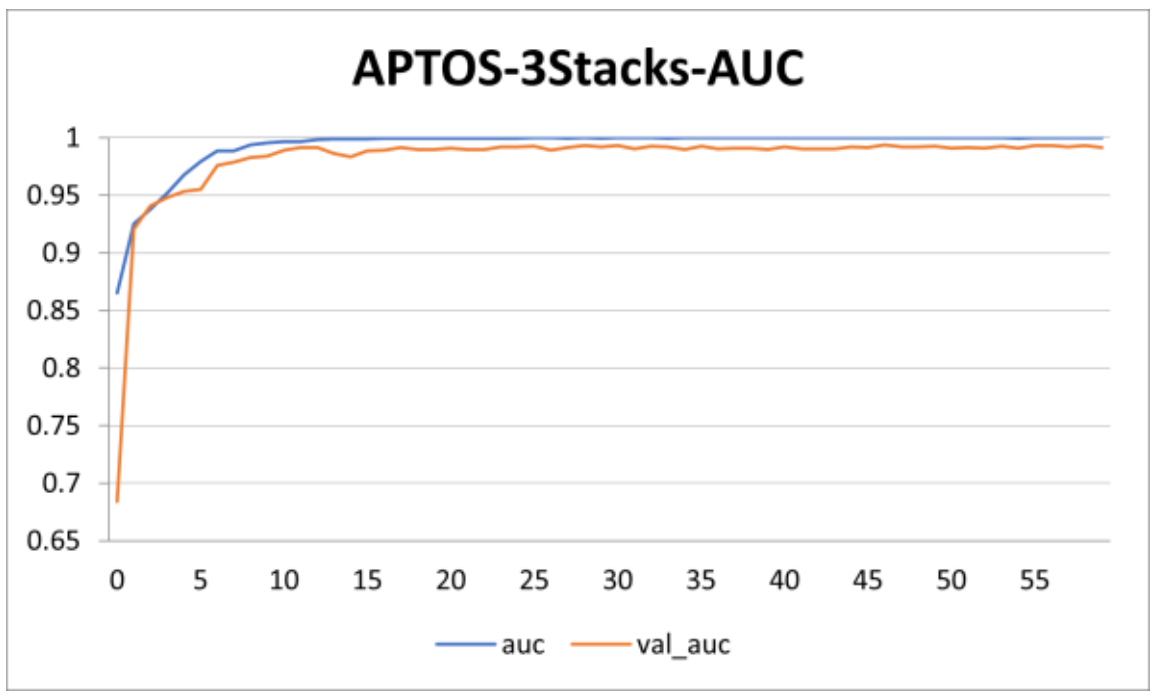

*Figure 15: APTOS-3Stacks-AUC*

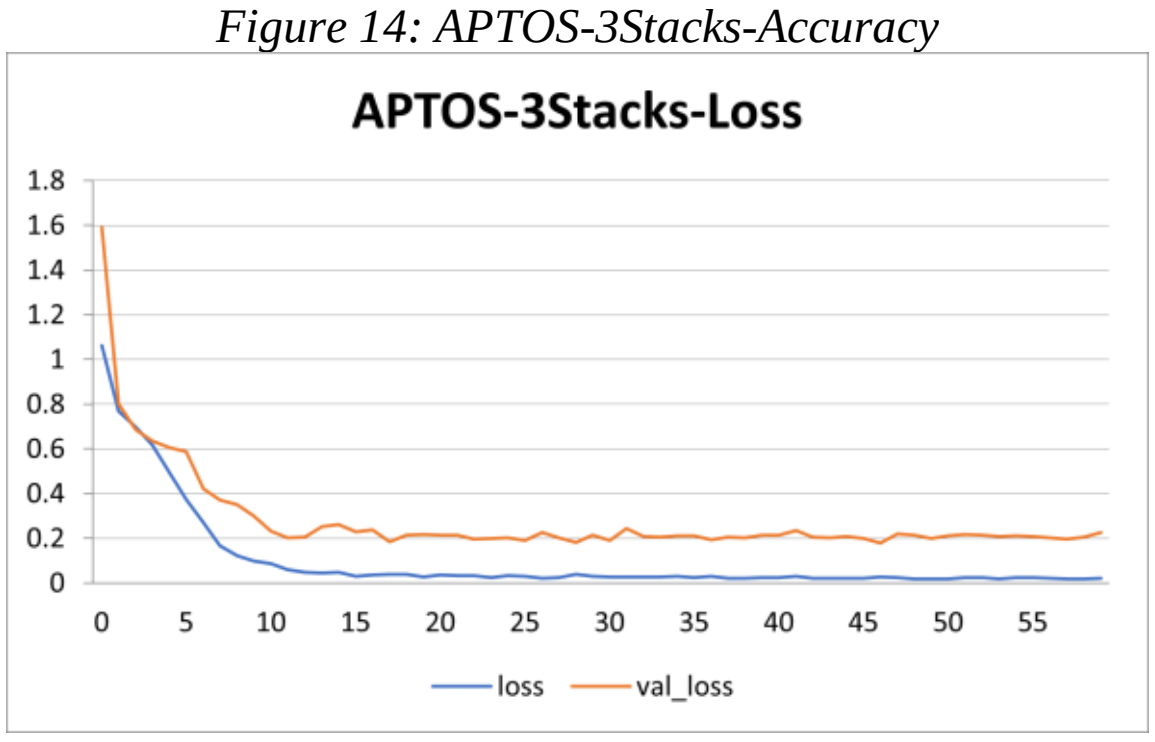

*Figure 16: APTOS-3Stacks-Loss*

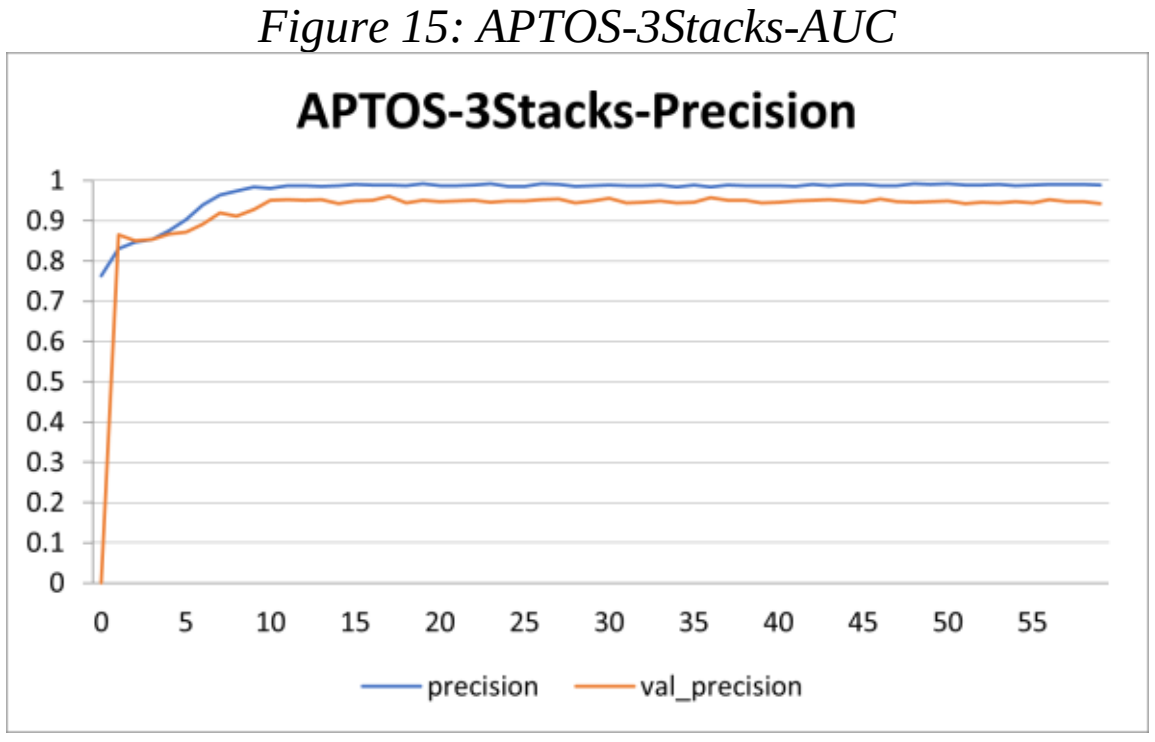

*Figure 17: APTOS-3Stacks-Precision*

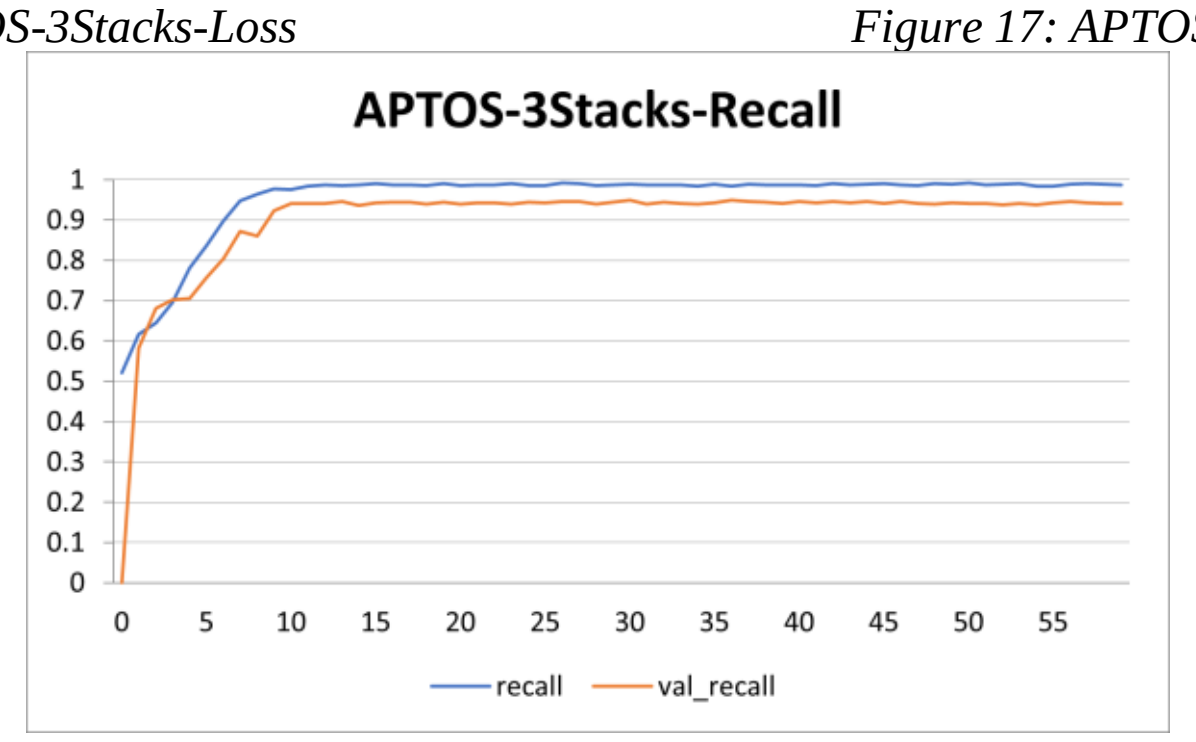

*Figure 18: APTOS-3Stacks-Recall*

For the APTOS-2019 Blindness Detection dataset, the models exhibited high training and validation accuracies across all stacks. Single-stacked U-Net achieved a training accuracy of 99.14% and a validation accuracy of 92.81%, with an AUC of 1.0000 for training and 0.9859 for validation. The loss values were 0.0168 during training and 0.3066 for validation, while precision and recall for validation stood at 93.38% and 92.03%, respectively. 2-Stacked U-Net demonstrated a training accuracy of 98.83% and a validation accuracy of 93.32%, maintaining an AUC close to perfection at 0.9999 for training and 0.9880 for validation. The associated loss values were 0.0187 for training and 0.2700 for validation, with precision at 93.80% and recall at 92.85%. 3-Stacked U-Net showed further improvement, achieving a training accuracy of 98.42% and a validation accuracy of 95.13%. The AUCs were 0.9999 and 0.9905 for training and validation, respectively, and the loss values were 0.0297 for training and 0.1924 for validation, with validation precision and recall recorded at 95.78% and 94.86%. Lastly, 4-Stacked U-Net reached a training accuracy of 99.11%, but dipped a little in validation accuracy at 92.13%, with an AUC of 0.9999 during training and 0.9816 for validation. The training and validation losses were 0.0221 and 0.3075, respectively, while precision and recall for validation were noted at 92.49% and 91.22%.

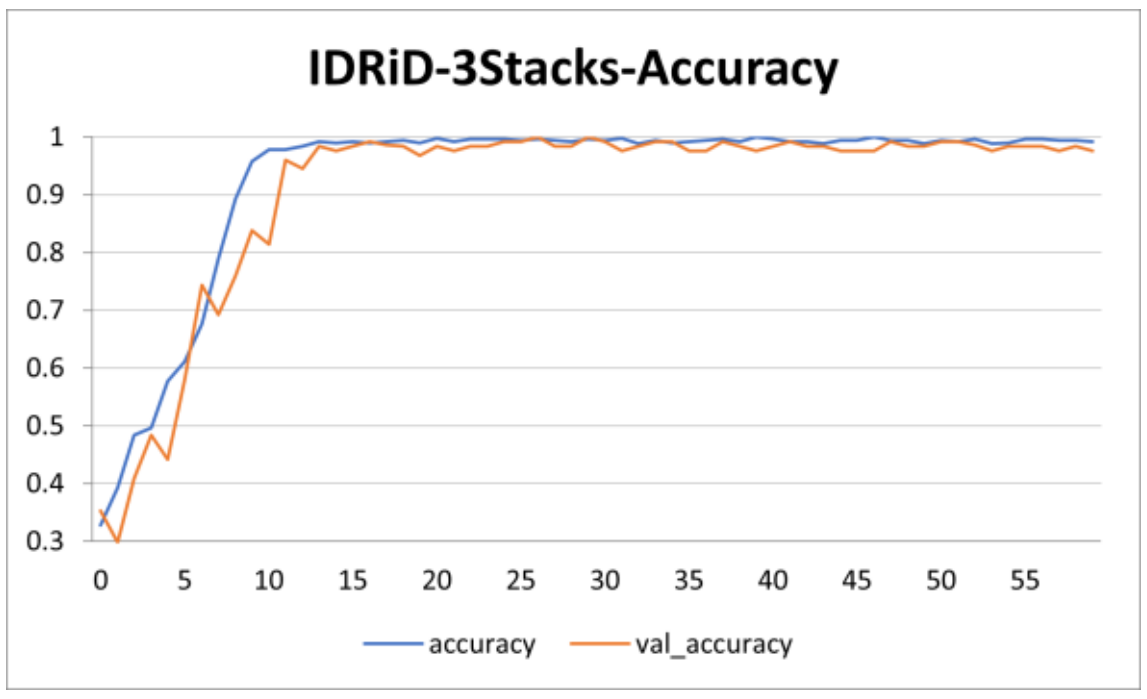

*Figure 19: IDRiD-3Stacks-Accuracy*

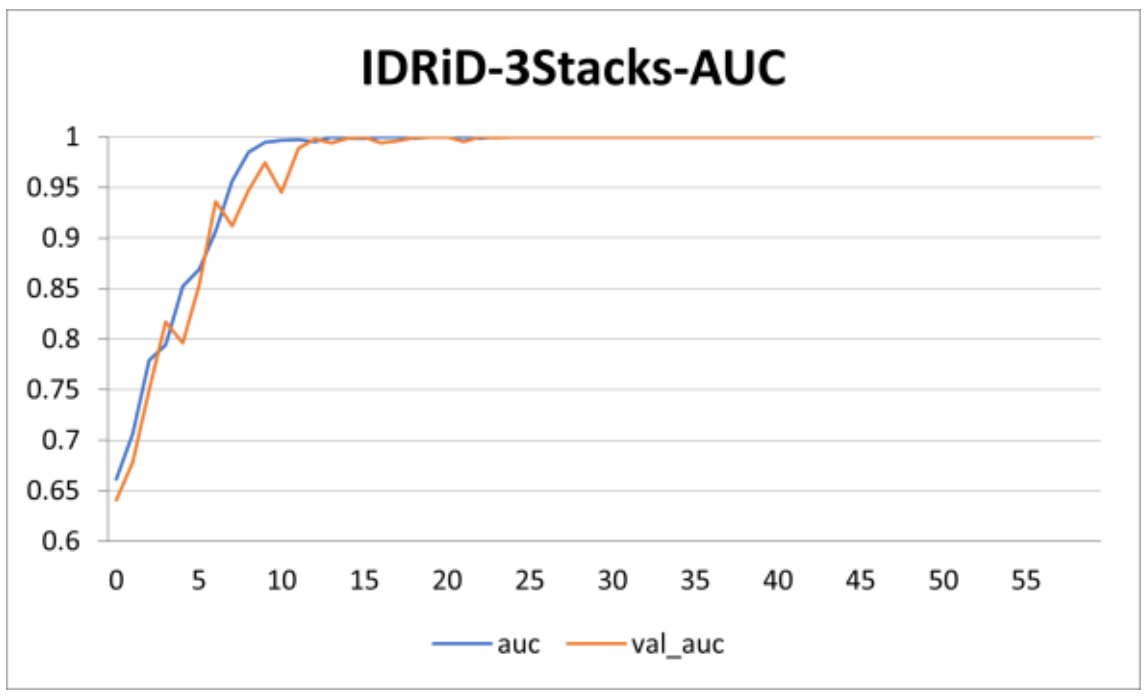

*Figure 20: IDRiD-3Stacks-AUC*

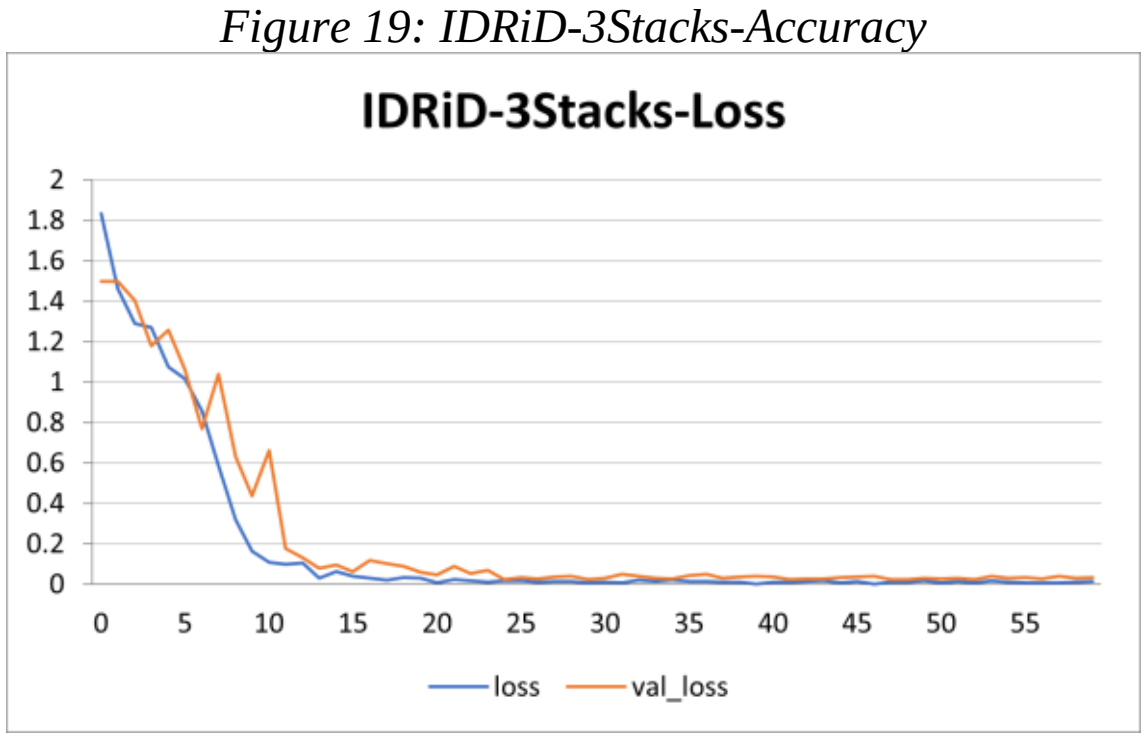

*Figure 21: IDRiD-3Stacks-Loss*

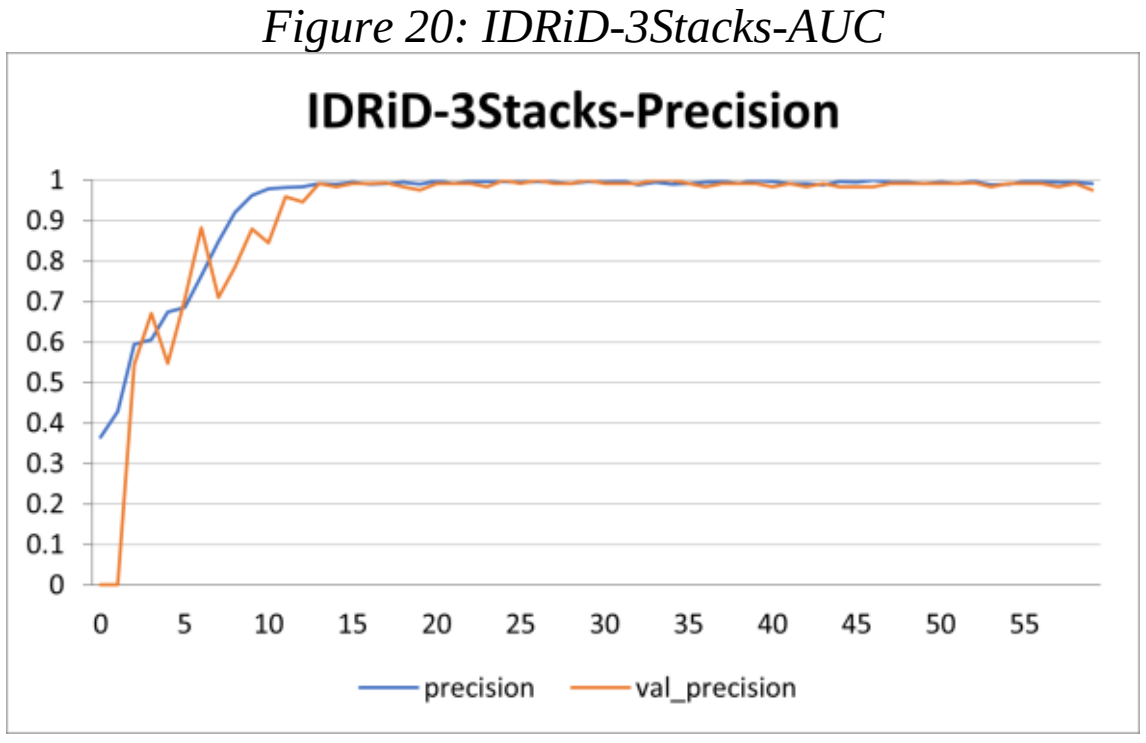

*Figure 22: IDRiD-3Stacks-Precision*

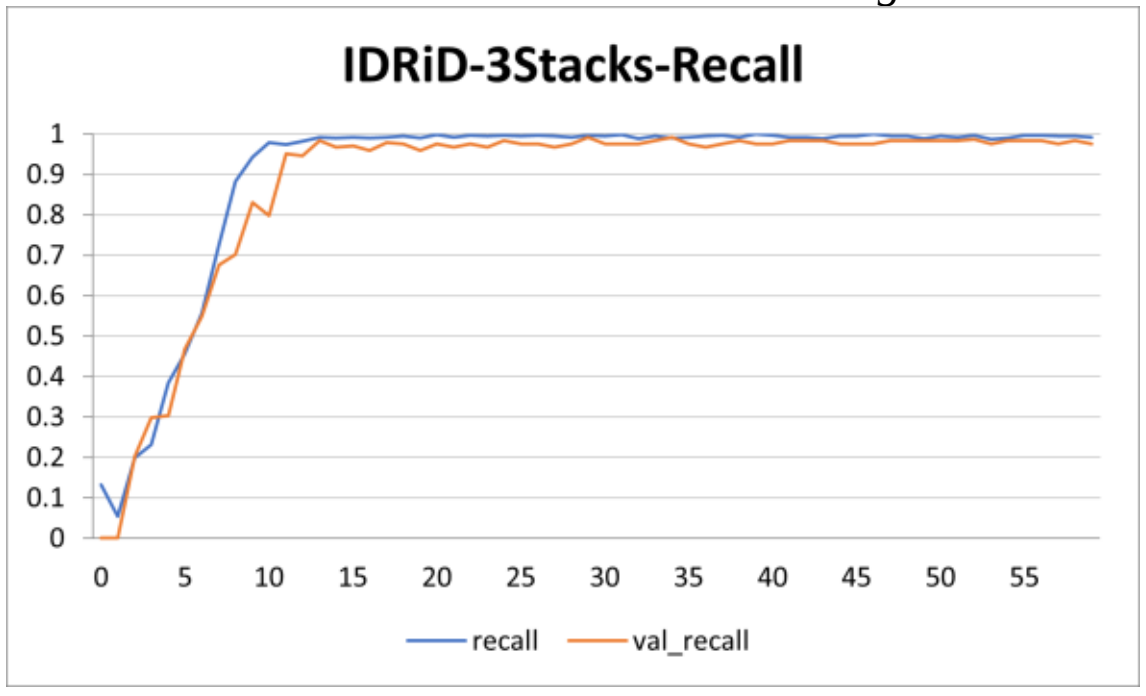

*Figure 23: IDRiD-3Stacks-Recall*

The Indian Diabetic Retinopathy dataset displayed even higher model performance. Single-stacked U-Net achieved an impressive training accuracy of 100% and a validation accuracy of 98.38%, with an AUC of 1.0000 for both metrics. The training loss was minimal at 0.0011, and the validation loss was 0.0370, with precision and recall for validation both at 98.38%. 2-Stacked U-Net demonstrated a training accuracy of 99.60% and a validation accuracy of 99.19%, maintaining an AUC of 1.0000 for training and 0.9998 for validation. The training loss was 0.0057, while the validation loss was 0.0347, with validation precision and recall reaching 99.19%. 3-Stacked U-Net also reported a perfect training accuracy of 100% and a validation accuracy flattened at 99.59%, with AUC scores at 1.0000 for both training and validation. The training and validation losses were recorded at 0.0072 and 0.0230, respectively, while validation precision reached 100%, with a recall of 99.19%. Stack 4 continued this high-performance trend with a training accuracy of 100% but a slightly lower validation accuracy of 99.19%. The AUC for training was 0.9999 and a perfect 1.0000 for validation, accompanied by training and validation losses of 0.0178 and 0.0062. Precision and recall for validation were both at 100%. Finally, 5-Stacked U-Net displayed a further dip in metrics compared to Stack-3, with a training accuracy of 99.80% and a validation accuracy of 95.95%, and an AUC of 1.0000 for training and 0.9996 for validation. The training loss was 0.0065, and the validation loss was 0.0960, with validation precision and recall at 95.95%.

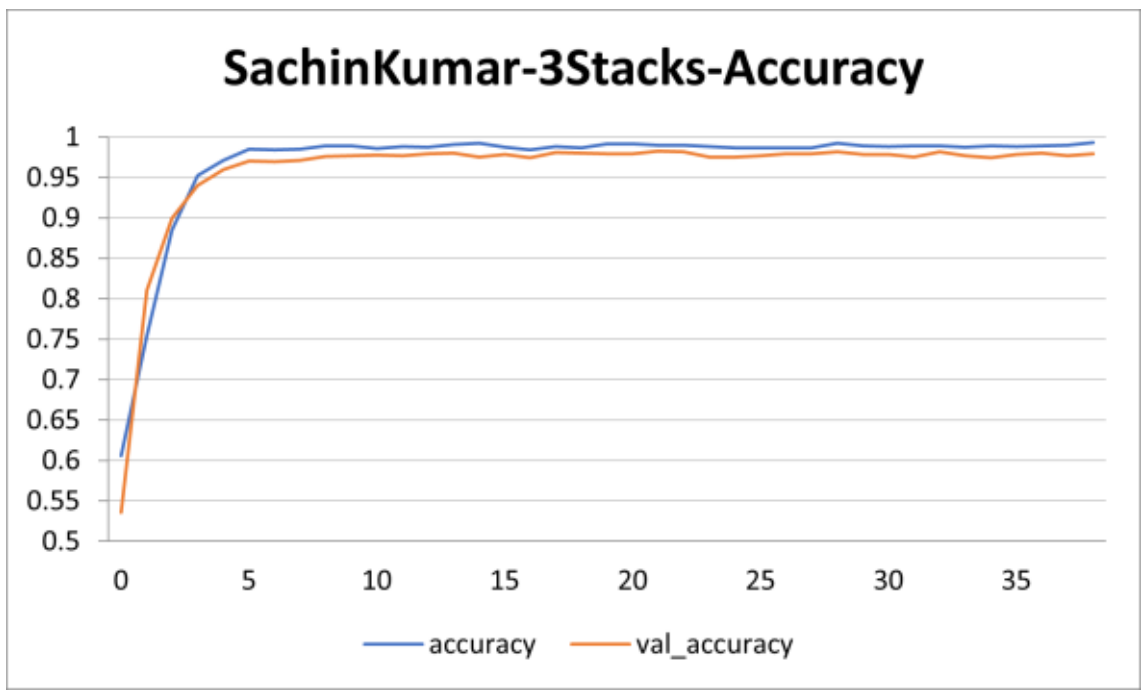

*Figure 24: SachinKumar-3Stacks-Accuracy*

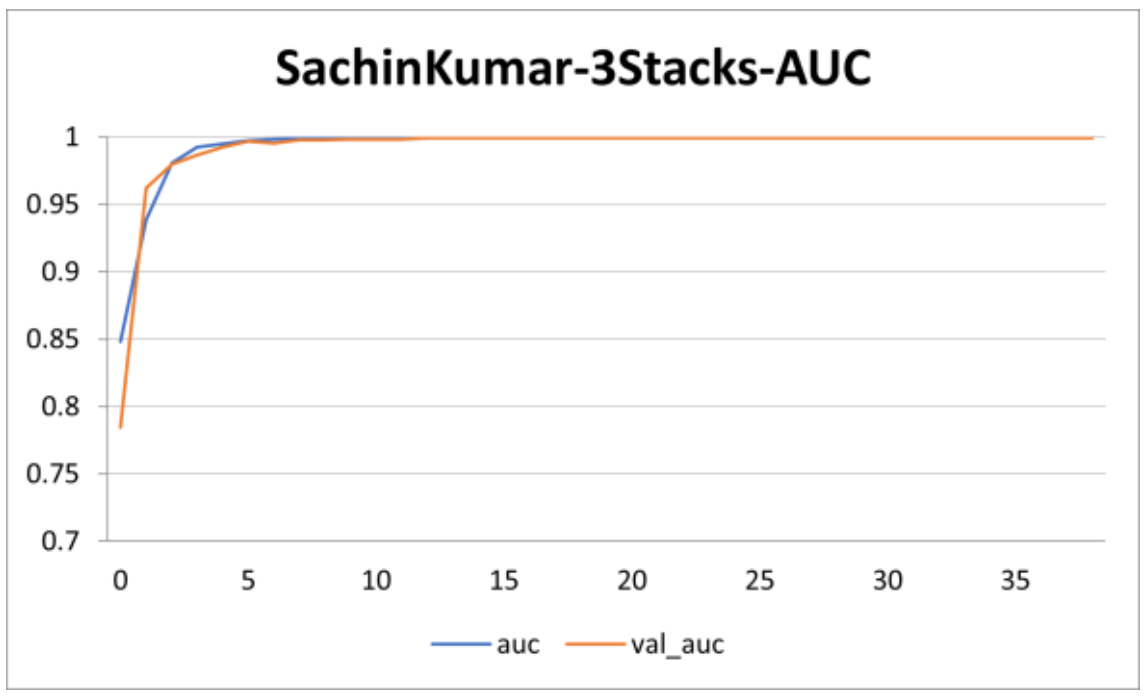

*Figure 25: SachinKumar-3Stacks-AUC*

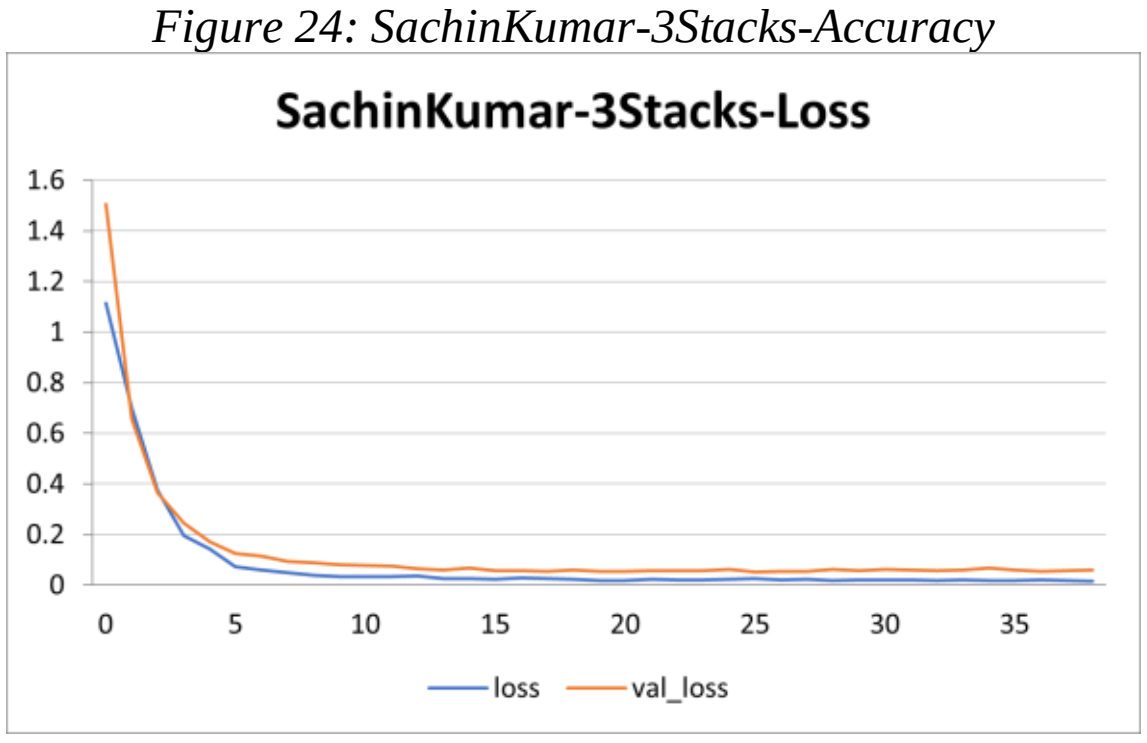

*Figure 26: SachinKumar-3Stacks-Loss*

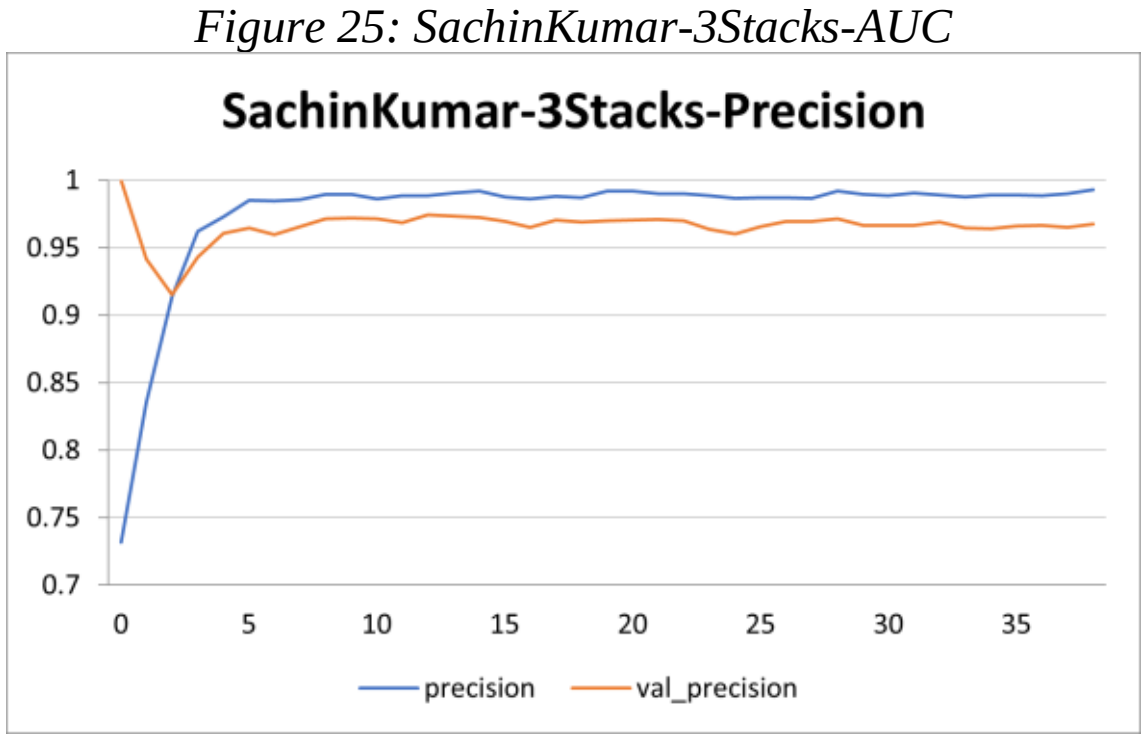

*Figure 27: SachinKumar-3Stacks-Precision*

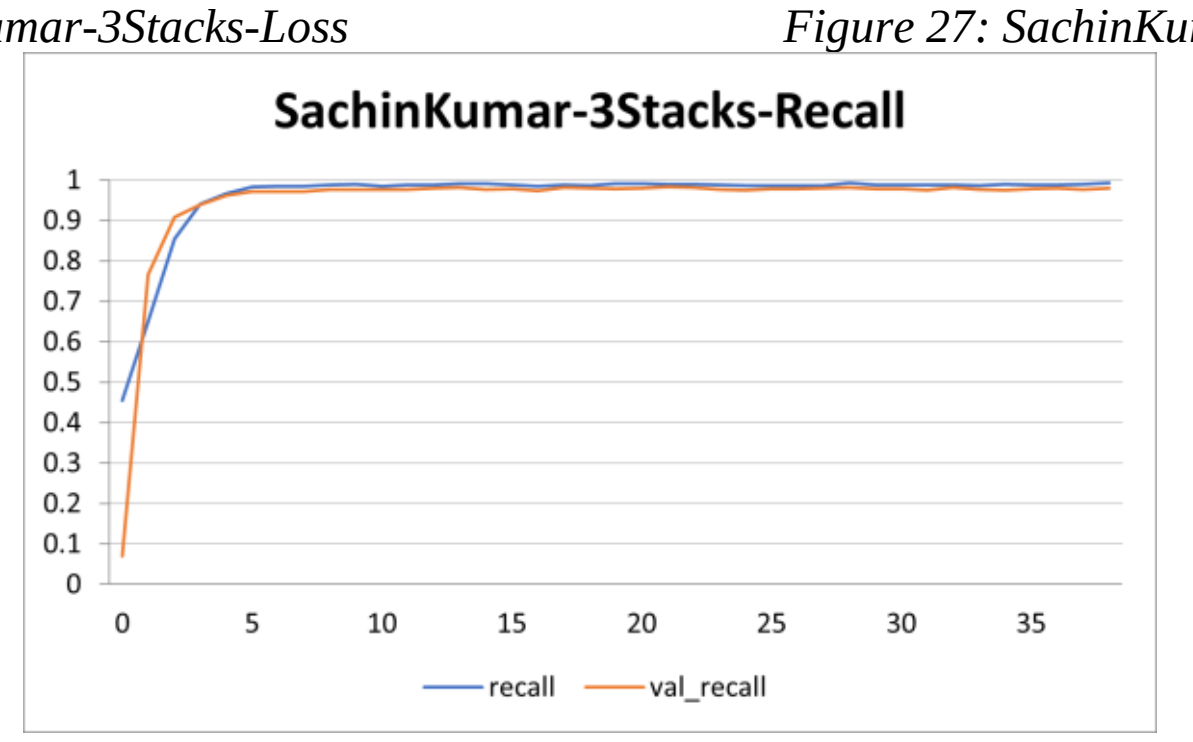

*Figure 28: SachinKumar-3Stacks-Recall*

For the Sachin Kumar Diabetic Retinopathy dataset, the results were consistently strong. The single-stacked U-Net exhibited a training accuracy of 99.15% and a validation accuracy of 97.83%, with an AUC of 1.0000 for training and 0.9997 for validation. Training and validation losses were 0.0152 and 0.0586, while precision and recall for validation were 97.94% and 97.71%, respectively. 2-Stacked U-Net reported a training accuracy of 99.04% and validation accuracy of 97.98%, with training and validation AUCs of 0.9999 and 0.9998. The training loss was 0.0165, and the validation loss was 0.0489, with validation precision and recall recorded at 98.22% and 97.86%. 3-Stacked U-Net achieved a peak training accuracy of 98.98% and a validation accuracy of 98.23%, with an AUC of 0.9999 for both training and validation. The training loss was 0.0223, and the validation loss was 0.0565, with precision and recall for validation noted at 97.07% and 98.21%. Lastly, the 4-Stacked U-Net demonstrated a training accuracy of 99.05% and a validation accuracy of 98.19%, with an AUC of 0.9999 for training and 0.9998 for validation. The training and validation losses were 0.0201 and 0.0455, while validation precision and recall were at 98.43% and 98.07%, respectively.

Among the various configurations tested, the model with three stacked layers consistently demonstrated superior performance across all datasets. It achieved the highest validation accuracy, AUC scores, and balanced precision and recall, indicating its robustness in detecting and classifying diabetic retinopathy. This suggests that the three-stacked UNET architecture is the most effective for our use case of detecting diabetic retinopathy, offering an optimal balance between model complexity and performance.

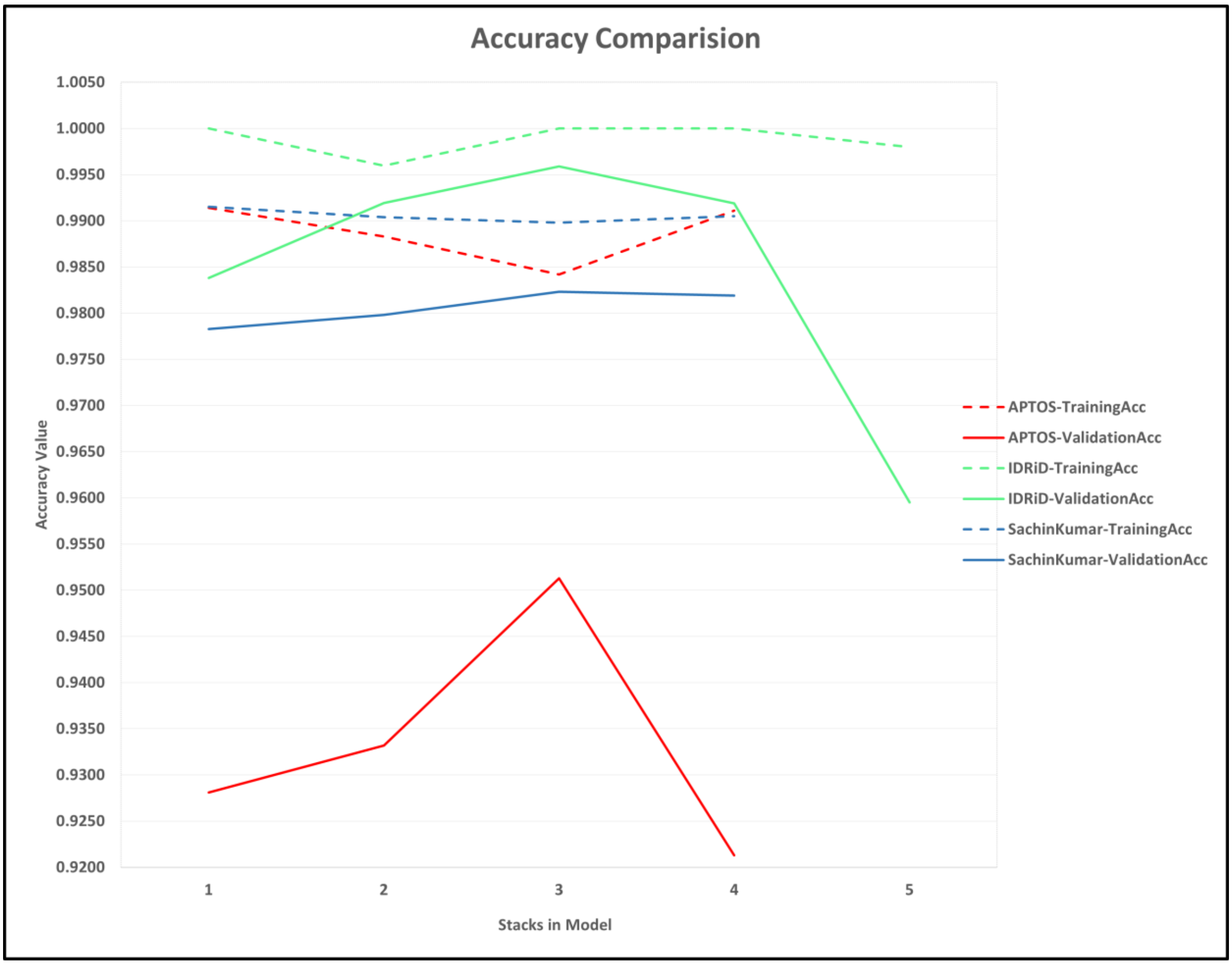

*Figure 29: Peak Accuracy Comparison*

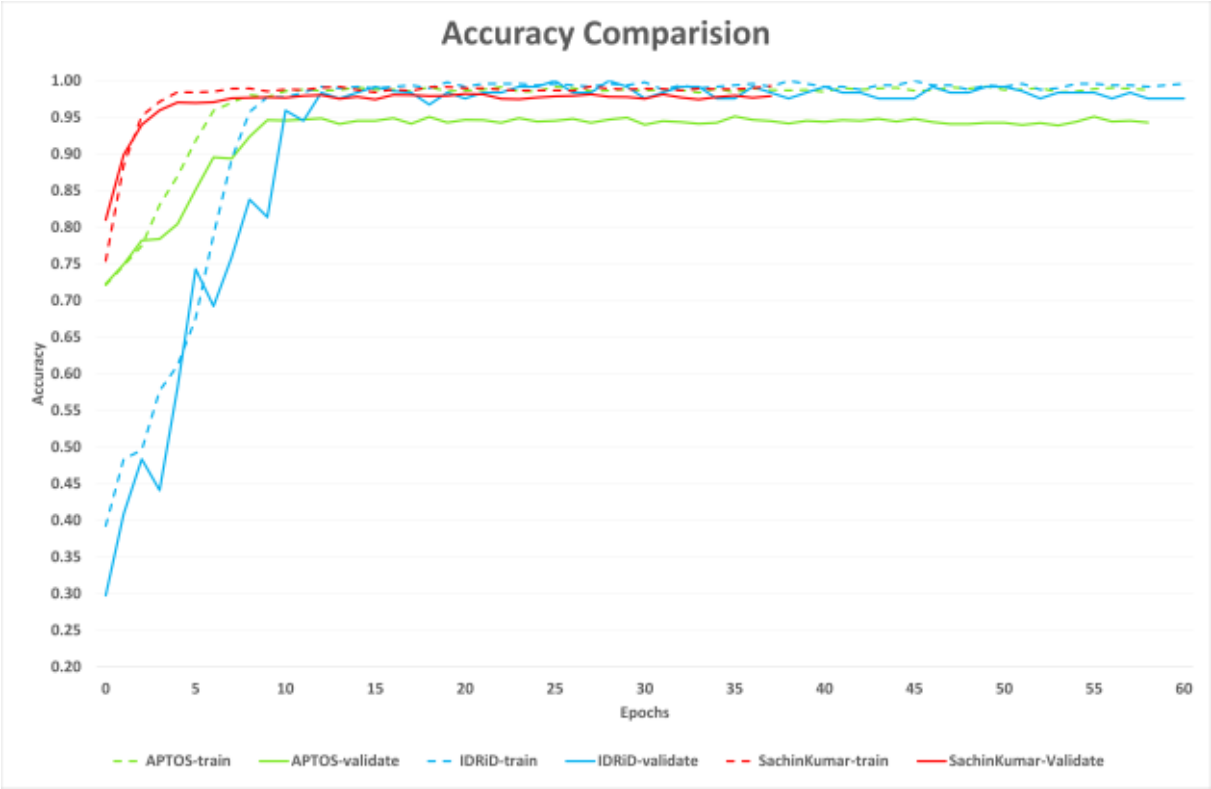

*Figure 30: 3rd Stack Accuracy Comparison*

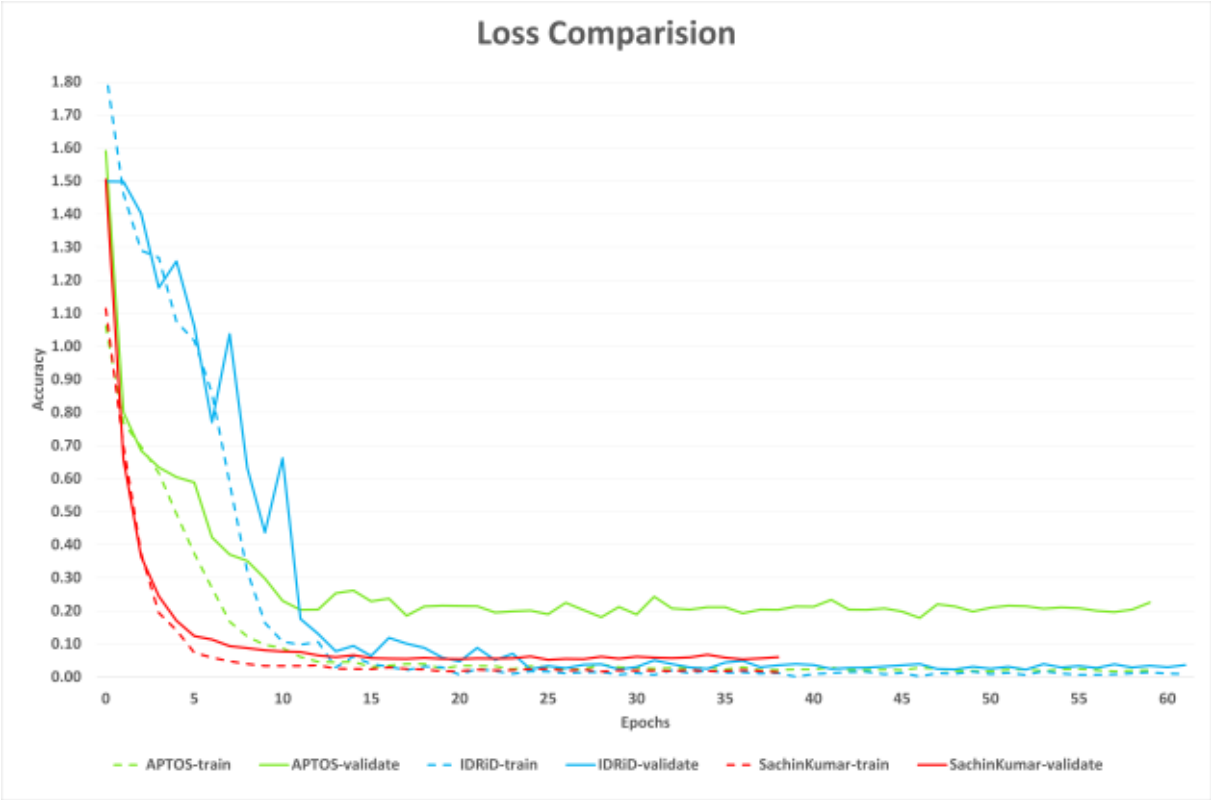

*Figure 31: 3rd Stack Loss Comparison*

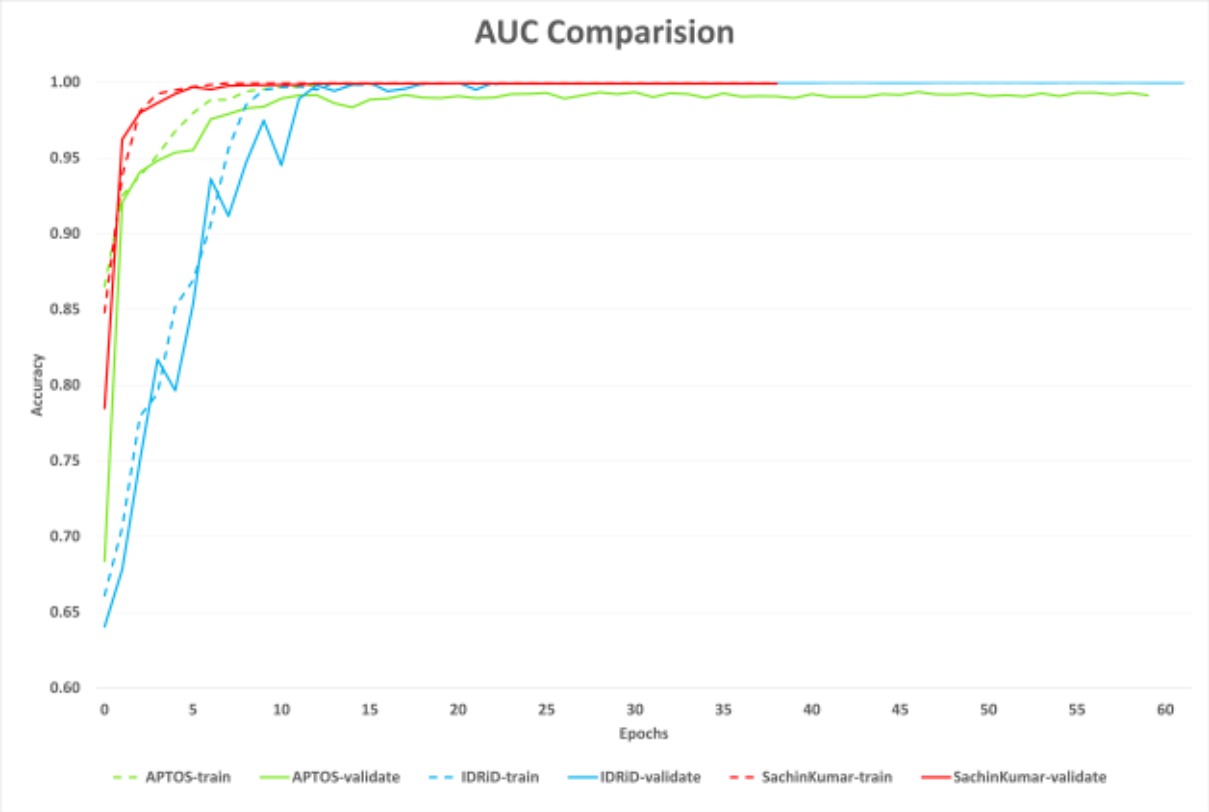

*Figure 32: 3rd Stack AUC Comparison*

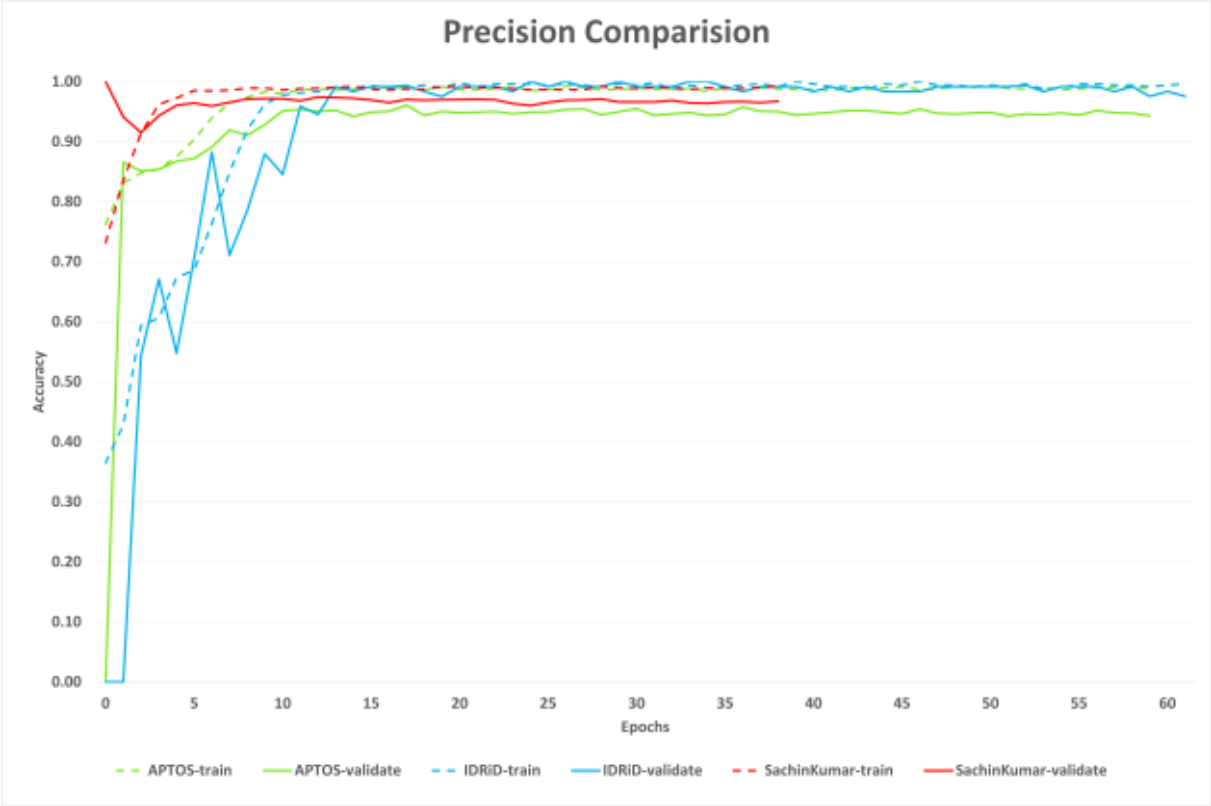

*Figure 33: 3rd Stack Precision Comparison*

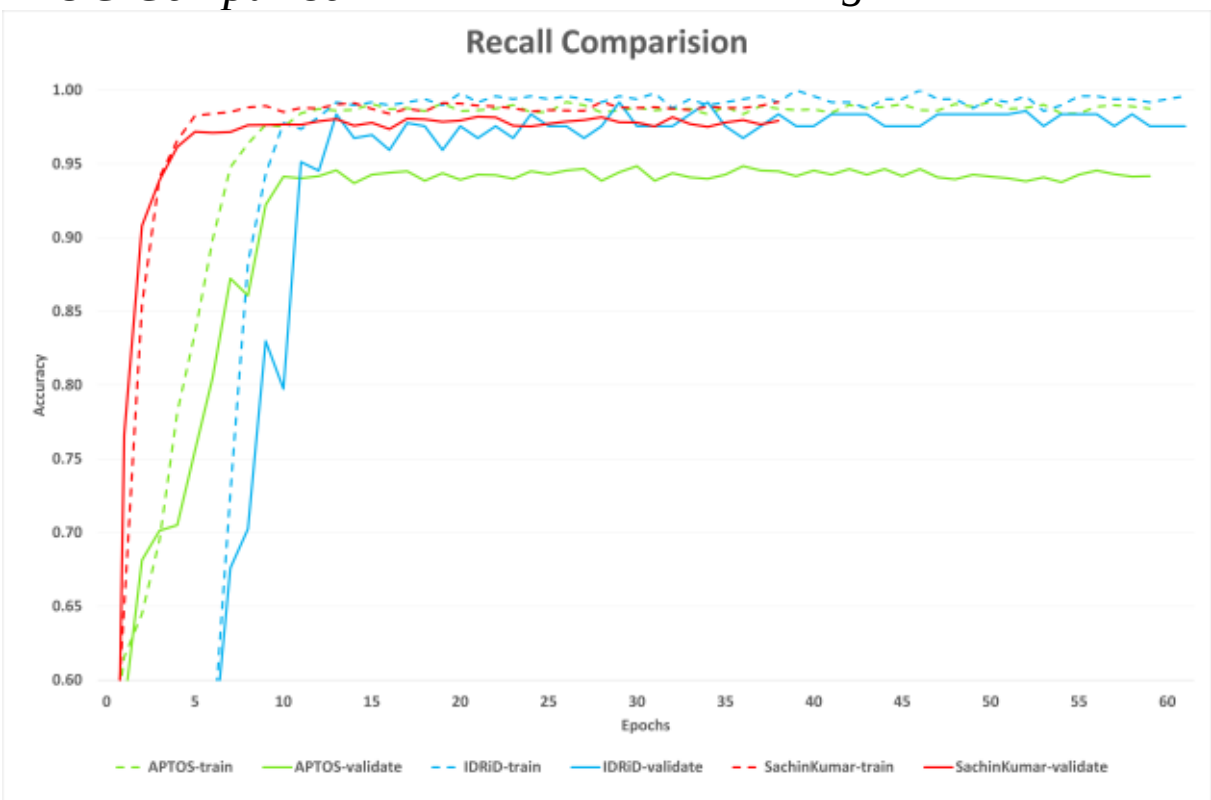

*Figure 34: 3rd Stack Recall Comparison*

# VI. CONCLUSION

This paper has demonstrated the effectiveness of the UNET and Stacked-UNET architecture, in detecting and classifying diabetic retinopathy (DR) from retinal fundus images. Through extensive experimentation with various stacking configurations, it was observed that the three-stacked UNET model consistently outperformed other configurations across multiple datasets.

We have reached a conclusion that 3-stacked UNET is the most optimal for the classification of fundus images into five stages of DR. Across all the three datasets we have achieved the highest level of metrics at the 3rd stack. In the IDRiD dataset the metrics of the fourth stack were on par with that of the third stack but dropped off at the 5th stack. Therefore, for computational simplicity, the third stack remains the best. The model is also able to deal with slightly worse quality of images as seen with the Diabetic Retinopathy dataset where the quality of the images slightly detoriates in comparison with the other two datasets.

These findings emphasize the robustness of the three-stacked UNET architecture, which provides an optimal balance between model complexity and performance. The consistent results across various datasets highlight its potential as a reliable tool for the early detection and classification of diabetic retinopathy. This study also emphasizes the importance of high-quality annotated datasets and effective data augmentation techniques to improve model training and generalization.

Future research could focus on integrating advanced data augmentation methods, utilizing larger and more diverse datasets, and expanding the model's application to other ophthalmic conditions. Further optimization of hyperparameters and exploration of ensemble learning approaches could also lead to improved performance.